\pdfoutput=1
\documentclass[11pt]{article}

\usepackage[preprint]{acl}

\usepackage{times}
\usepackage{latexsym}
\usepackage{algorithm} 
\usepackage{algpseudocode} 
\usepackage{amsmath}
\usepackage{xcolor}
\usepackage{tikz}
\usepackage{soul}
\usepackage{ctable}
\usepackage{amsmath}
\usepackage[most]{tcolorbox}
\usepackage{amssymb}
\usepackage{enumerate}
\usepackage{enumitem}
\usepackage{siunitx} 
\usepackage{xurl} 
\usepackage{graphicx}
\usepackage{epstopdf}

\usepackage{bbm}

\usepackage[T1]{fontenc}

\usepackage[utf8]{inputenc}

\usepackage{microtype}

\usepackage{inconsolata}

\usepackage{graphicx}

\usepackage[table]{xcolor}

\definecolor{lightred}{RGB}{255,230,230}
\definecolor{lightgreen}{RGB}{230,255,230}
\usepackage[table]{xcolor}

\newcommand{\sigA}[1]{\cellcolor{sigone}#1\textsuperscript{*}}
\newcommand{\sigB}[1]{\cellcolor{sigtwo}#1\textsuperscript{**}}
\newcommand{\sigC}[1]{\cellcolor{sigthree}#1\textsuperscript{***}}

\definecolor{sigone}{RGB}{255,250,230}
\definecolor{sigtwo}{RGB}{255,238,200}
\definecolor{sigthree}{RGB}{255,220,170}

\usepackage[table]{xcolor}
\definecolor{poscoef}{RGB}{230,245,255}
\definecolor{negcoef}{RGB}{255,235,235}

\newcommand{\toppos}[1]{\cellcolor{poscoef}\textbf{#1}}
\newcommand{\topneg}[1]{\cellcolor{negcoef}\textbf{#1}}

\newcommand{\cohedown}[1]{\cellcolor{lightred}{\scriptsize$\downarrow$#1}}
\newcommand{\coheup}[1]{\cellcolor{lightgreen}{\scriptsize$\uparrow$#1}}

\newcommand{\dissdown}[1]{\cellcolor{lightgreen}{\scriptsize$\downarrow$#1}}
\newcommand{\dissup}[1]{\cellcolor{lightred}{\scriptsize$\uparrow$#1}}

\title{SARA: Stress Test Reasoning in Audio Deepfake Detection}

\author{
 \textbf{Binh Nguyen\textsuperscript{1}}\;\;\;\;
 \textbf{Charles Fleming\textsuperscript{2}}\;\;\;\;
 \textbf{Thai Le\textsuperscript{1}}
 \\
 \textsuperscript{1}Indiana University\;\;\;\;
 \textsuperscript{2}Cisco Research\;\;\;\;
\\
   \textsuperscript{1}\texttt{\{binhnguy,tle\}@iu.edu}\;\;\; 
   \textsuperscript{2}\texttt{chflemin@cisco.com}
}

\begin{document}
\maketitle
\begin{abstract}
Audio Language Models (ALMs) offer a promising shift towards explainable audio deepfake detections (ADD), moving beyond \textit{black-box} classifiers by providing transparency to their predictions via reasoning traces. However, such reasoning may not support the model predictions, reflecting poor coherence, or, worse, may rationalize incorrect predictions with plausible but misleading explanation. Moreover, the behavior of ALM reasoning under adversarial attacks remains under-explored, raising questions about the practical reliability of such explanation capabilities. To address this gap, this study introduces \textbf{SARA} (\textbf{S}hift \textbf{A}nalysis of \textbf{R}easoning in \textbf{A}udio), a diagnostic framework that evaluates ALM reasoning across three dimensions: acoustic perception, reasoning-verdict coherence and dissonance. We test five open-source ALMs against both acoustic and linguistic adversarial attacks. We show that acoustic attacks significantly degrade reasoning-verdict coherence (average decrease of 14.20\%), frequently inducing internal logical conflicts. Conversely, linguistic attacks achieve higher attack success rates while maintaining reasoning coherence. We further demonstrate that the textual coherence of generated reasoning traces also serves as a latent indicator of adversarial inputs, enabling effective detection of perturbed audio (0.78 in F1) \textit{without accessing the raw acoustic signal}. These findings suggest that reasoning traces provide diagnostic utility that persists even when final classification outputs are compromised.

\end{abstract}

\section{Introduction}

\begin{figure}[tb!]
    \centering
    \setlength{\fboxsep}{0pt}
    \includegraphics[width=7.5cm]{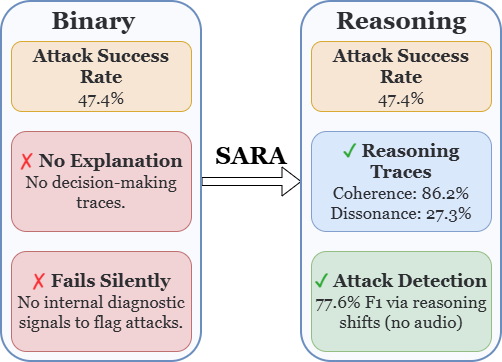}
    \caption{\textbf{SARA} bridges the gap between binary detectors and explainable ADD. While traditional classifiers fail silently under attack, SARA leverages ALM reasoning shifts (coherence and dissonance) to detect perturbations even when final verdicts are compromised.}
    \label{fig:motivation}
    \vspace{-15pt}
\end{figure}

In high-stakes domains such as forensic audio analysis and legal proceedings, standard binary deepfake detectors, which classify audio as "real" or "fake", are increasingly insufficient. For automated detection outputs to be admissible under rigorous legal standards like the amended US Federal Rule of Evidence 702 \cite{pusey2026rethinking} or to satisfy the strict transparency and explainability mandates of international frameworks like the EU AI Act \cite{eu_ai_act_2024}, forensic experts require auditable, step-by-step explanations. This operational necessity has driven a paradigm shift toward reasoning-capable Audio Language Models (ALMs). These models provide a "glass-box" view of their decisions by generating intermediate explanations to support their final classification verdicts. 


Prior research has extensively explored the interpretability of ADD decisions; however, existing approaches primarily rely on post-hoc attribution methods such as Occlusion and Attention Visualization \cite{channing2024robustrealworldaudiodeepfake}, Segmental Speech Features \cite{Yang_2026}, and Temporal Class Activation \cite{Li_2024}. In contrast, frontier multimodal models (e.g., GPT-4o or Gemini 3.1 Pro) generate natural language reasoning steps, theoretically enabling humans to verify the model's intermediate logic. Building upon this direction, recent work has begun applying open-source, reasoning-capable ALMs to ADD tasks, including models such as Phi-4-multimodal \cite{microsoft2025phi4minitechnicalreportcompact}, granite-speech \cite{saon2025granitespeechopensourcespeechawarellms}, Qwen2-Audio \cite{chu2024qwen2audiotechnicalreport}, and gemma-3n-E4B \cite{gemmateam2025gemma3technicalreport}, where step-by-step explanations support the final classification verdict.

A secondary advantage of reasoning ALMs lies in their potential diagnostic utility, which may persist even when the final classification fails under adversarial conditions. Traditional binary ADD systems are known to collapse under acoustic perturbations, such as background noise or time-stretching \cite{Kawa2022DefenseAA, uddin2025adversarialattacksaudiodeepfake}, and they remain vulnerable to subtle linguistic variations \cite{nguyen-binh-2025-turing, nguyen-etal-2025-read}. Such vulnerabilities render binary labels insufficient for high-stakes environments. Specifically, as highlighted by \cite{xie2025fakesound2benchmarkexplainablegeneralizable}, black-box systems typically lack the capacity to: (1) localize forgery timestamps; (2) distinguish between specific manipulation techniques; or (3) trace the provenance of synthetic content.

Driven by the need for verifiable decision-making, this study evaluates the robustness of integrating ALMs with explicit Chain-of-Thought (CoT) \cite{wei2023chainofthoughtpromptingelicitsreasoning} reasoning for ADD. This integration shifts the inquiry from a binary ``Is it fake?'' to the forensically critical ``Why is it fake?''. To systematically assess whether ALMs can serve as reliable explainable systems, we introduce \textbf{SARA}, a three-tier diagnostic evaluation framework. We investigate the robustness of reasoning through three key \textbf{r}esearch \textbf{q}uestions: \textbf{RQ1} (Acoustic Perception): Does the model accurately perceive the raw audio signal?  \textbf{RQ2} (Reasoning-Verdict Coherence): Do the generated intermediate reasoning steps entail the final predicted label? and  \textbf{RQ3} (Reasoning-Verdict Dissonance): Does the reasoning trace exhibit internal conflict that flags an anomaly even when the classification head is deceived by adversarial attacks? Our findings are as follows:

\begin{enumerate}[leftmargin=\dimexpr\parindent-0\labelwidth\relax,noitemsep,topsep=0pt]

\item \textbf{Modality-Specific Patterns of Adversarial Degradation:} We identify that acoustic and linguistic perturbations induce fundamentally different statistical shifts in model reasoning. Acoustic attacks cause significant internal incoherency, resulting in an average coherence degradation of 14.20\% ($p {<} 0.0001$) and measurable increases in reasoning-verdict dissonance (e.g., up to a +19.4\% shift in Phi-4-multimodal). Conversely, linguistic perturbations achieve a higher aggregate Attack Success Rate (61.73\% vs. 47.38\% for acoustic attacks) while inducing minimal coherence erosion (a 3.91\% average drop). This indicates that linguistic attacks systematically induce classification errors without generating detectable statistical anomalies in the reasoning trace.

\item \textbf{Adversarial Detection via Reasoning Shifts:} We show that statistical shifts within intermediate reasoning can serve as latent indicators of adversarial perturbations, even without analyzing the audio. By training a binary classifier on reasoning coherence and perception metrics, we achieve up to a 77.59\% F1 score in detecting perturbed audio. Regression analysis indicates that when a model’s reasoning regarding Prosody erodes from coherent to incoherent, the probability of an adversarial attack increases by 6.07\%.
\end{enumerate}

\section{Related Works}
\textbf{Audio Deepfake Detection (ADD)} research has primarily focused on customizing neural architectures for binary classification. RawNet-2~\cite{tak2021endtoendantispoofingrawnet2} shifted to raw waveforms, employing distinct filter banks for discriminative cues without handcrafted features. AASIST-2~\cite{tak2022automaticspeakerverificationspoofing} utilized graph attention networks to model complex spectral-temporal dependencies, setting a high-performance benchmark. CLAD \cite{wu2024cladrobustaudiodeepfake} introduced learning objectives to enhance robustness against diverse acoustic conditions and unseen attacks. More recently, ALLM4ADD \cite{10.1145/3746027.3755851} explored the application of ALMs to ADD tasks, though it does not explicitly address the models' reasoning capabilities or provide an in-depth analysis of their robustness.

\noindent \textbf{Explainability} in ADD has largely relied on post-hoc methods for non-ALM detectors, including occlusion sensitivity and attention roll-out for identifying influential spectral bands or temporal frames~\cite{channing2024robustrealworldaudiodeepfake}, temporal localization of manipulated segments~\cite{Yan2024temporal, xie2023efficienttemporarydeepfakelocation}, and SHAP-based attribution of spectrogram artifacts~\cite{Ge2021ExplainingDL}. In contrast, ALMs offer a more natural form of explanation through textual reasoning traces, which are more user-friendly and directly interpretable to end users.


\noindent \textbf{Audio Language Models.} Benchmarks for ALMs evaluate acoustic reasoning, requiring models to analyze audio content based on natural language instructions. AIR-Bench \cite{yang2024airbenchbenchmarkinglargeaudiolanguage} covers four dimensions: speech, sound, music, and mixed audio. MMAU \cite{ICLR2025_d36f2089} is a dataset evaluating audio-based understanding and reasoning via multiple-choice questions. SpeechR \cite{yang2025speechrbenchmarkspeechreasoning} is a benchmark rigorously testing factual, procedural, and normative reasoning in spoken interactions.

\section{Proposed SARA Framework} \label{sec:formulation}
\subsection{Formulation}
Let $X{\in} \mathbb{R}^{L}$ be the input audio with wavelength $L$. We define ALMs with CoT as $\mathcal{F}(X){\rightarrow}Y$ mapping $X$ to generate the multi-aspect reasoning verdict $Y{=}\{r_1,r_2,\ldots,r_N, v\}$. Here, $v{\in}\{\text{fake}, \text{real}\}$ is the final verdict, and each text $r_i$ represents a reasoning corresponding to the $i^{th}$ aspect. To align the model's reasoning process with human intuition, we adopt the reasoning taxonomy for ADD established by \cite{10.1145/3658644.3670325}. This taxonomy was established by analyzing 24,240 open-text responses from 1,212 participants, which were independently evaluated and reconciled by three raters to achieve strong consensus ($\kappa = 0.82$). Consequently, there are six distinct reasoning dimensions:
\begin{itemize}[leftmargin=\dimexpr\parindent-0.2\labelwidth\relax,noitemsep,topsep=0pt]
    \item \textbf{Prosody:} Tone, pitch, inflections, and emotion. \cite{10.1145/3658644.3670325} identified prosody as the most common linguistic factor humans use, looking for ``robotic'' flatness or unnatural cadence.
    \item \textbf{Disfluency:} Presence of natural imperfections such as ``um'', ``uh'' fillers, hesitations, and stuttering, which are often absent in synthesized speech.
    \item \textbf{Speed:} Pacing of speech to detect unnatural rushing or dragging that signifies algorithmic generation.
    \item \textbf{Speaking Style:} Articulation, accents, and dialect consistency. This captures whether the voice sounds like ``scripted speech'' versus spontaneous conversation.
    \item \textbf{Liveliness:} Presence of breathing sounds, mouth noises, and nasal intake. The absence of these ``signs of life'' is a strong indicator of synthetic audio.
    \item \textbf{Quality:} Environmental and technical artifacts, such as background noise, static, clipping, or the ``sterile'' silence typical of generated audio.
\end{itemize}

Our goal is to comprehensively analyze the \textbf{reasoning pathways} of ALMs by posing several fundamental questions about their decision-making logic. 
First, do ALMs actually perceive acoustic properties similarly to humans, such as detecting natural breathing pauses or identifying raw microphone feedback. What if they achieve high classification accuracy despite having \textbf{weak perceptual grounding}? 
If a model does accurately perceive these forensic features, how does it curate these raw observations to construct a coherent \textbf{reasoning-verdict explanation}?
Furthermore, just as human reasoning struggles under pressure, how do ALMs behave under \textbf{adversarial strain}? 
Specifically, does the connection between their internal reasoning steps and final verdicts become decoupled? 
If the final classification is successfully bypassed, does this internal conflict leave detectable traces of the stress-testing? 
Crucially, can these distinct shifts in the reasoning trace, such as a spike in \textbf{internal dissonance} when the model incorrectly classifies a corrupted audio sample, be leveraged to identify adversarial manipulation even when the model's final verdict fails?
To stress test ALM reasoning, we define an \textbf{adversarial input} $\tilde{X}{=}Adv(X, \theta)$, where $Adv{\in}\{\text{linguistic attack},  \text{acoustic attack}\}$ and $\theta$ are corresponding attack hyper-parameters (e.g., noise SNR, pitch variance, or voice profiles).

\subsection{Acoustic Perception (RQ1)}
\label{sec:rq1_perception}

Before analyzing \textit{how} the model structures its arguments, we first determine whether its reasoning is grounded in the actual physical signal. We introduce an \textbf{Acoustic Perception} benchmark to evaluate the accuracy of its underlying acoustic perception. To quantify this, we curate a benchmark set $\mathcal{D}_{Perc}$ by aggregating expert-annotated samples from two existing datasets: MMSU \cite{wang2026mmsu} and MMAU-Pro \cite{kumar2025mmauprochallengingcomprehensivebenchmark}. We organized these samples into the six key reasoning aspects ($\S$\ref{sec:formulation}).
We define a probabilistic \textbf{verification} function $\mathcal{V} (X, q)$ that answers : ``\textit{Does the model's response to question $q$ match the ground-truth attribute of audio $X$?}'' (1 for match, 0 for mismatch). 
Operationally, we prompt the ALM with the audio and question, validate its generated answer against the ground truth. For the $i^{th}$ reasoning aspect $r_i$ (e.g., Liveliness) associated with a bank of questions $\mathcal{Q}_i$, \textbf{perception} score represents the overall, average accuracy:
\begin{equation}
\label{eq:perception_score}
\Phi_{\text{Perc}}(r_i)
=
\mathbb{E}_{X \sim \mathcal{D}, \, q \sim \mathcal{Q}_i}
\left[
\mathcal{V}(X, q)
\right],
\end{equation}
\noindent where a high perception ($\Phi_{\text{Perc}}{\approx}1$) indicates the model perceives the audio features correctly.

\subsection{Reasoning-Verdict Coherence (RQ2)}
\label{sec:rq2_coherence}
After establishing acoustic grounding, we can now evaluate the coherence of its textual explanations. We define \textbf{Reasoning-Verdict Coherence} as the degree of textual entailment between the intermediate reasoning steps (premises) and the final classification verdict (hypothesis). In forensic applications, an explanation is reliable if these premises support the verdict. For instance, if a model reasons ``unnatural robotic artifacts'' but predicts ``real'', it exhibits an entailment failure. Hence, even an ALM with perfectly accurate acoustic perception can be incoherent with unreliable verdicts.

To quantify this coherency, we define a probabilistic \textbf{entailment} function $\mathcal{E} (r_i, v)$ that classifies: ``Does the reasoning aspect $r_i$ entails the final verdict $v$?'' (1 for yes, 0 for no) - details in Appendix \ref{app:imp_entailment}. For an aspect $r_i$, this score represents the estimated likelihood that the explanation supports its own conclusion $v$, \textit{regardless} of whether that verdict is factually correct:
\begin{equation}
\label{eq:coherence_score}
\Phi_{\text{Coh}}(r_i)
=
\mathbb{E}_{v \sim \mathcal{D}}
\left[
\mathcal{E}(r_i, v)
\right], 
\end{equation}
\noindent where a high coherence ($\Phi_{\text{Coh}}{\approx}1$) indicates the model constructs coherent reasoning-verdict entailments.

\paragraph{Coherence under Attacks.}
While $\Phi_{\text{Coh}}$ measures baseline coherence, stress-testing robustness requires observing how this consistency holds under adversarial pressure. SARA quantifies this via \textbf{coherence shift} ($\Delta\Phi$) between Original (ORG) and Perturbed (PER) inputs:
\begin{equation}
\label{eq:coherence_shift}
\Delta\Phi_{\text{Coh}}(r_i) = \Phi_{\text{Coh}}^{\text{PER}}(r_i) - \Phi_{\text{Coh}}^{\text{ORG}}(r_i),
\end{equation}
This metric quantifies the degree to which an attack impacts the model's coherency. A positive shift ($\Delta\Phi \ge 0$) indicates the reasoning-verdict link remains coherence despite the perturbation. Conversely, a decline ($\Delta\Phi \ll 0$) signals \textit{coherence erosion}, where model's coherency is prone to errors under perturbation.

However, changes in coherence alone do not identify the type of failure caused by an attack. If the attack changes the model's final prediction, often indicated by a high \textbf{A}ttack \textbf{S}uccess \textbf{R}ate (ASR), a small change in $\Phi_{\text{Coh}}$ can indicate that the model still produces a coherent explanation, but now for an incorrect verdict. This is a concerning failure mode because the reasoning trace may make the wrong decision appear justified. In contrast, when the attack does not change the final prediction, a decrease in $\Phi_{\text{Coh}}$ may indicate that the perturbation disrupts the model's reasoning process even though the final verdict remains correct. Such internal inconsistency can provide a useful signal for detecting adversarial inputs. Hence, to distinguish between models that are coherently wrong and models that reveal uncertainty or conflict in their reasoning, we analyze reasoning traces specifically where the \textit{final classification fails}.

\subsection{Reasoning-Verdict Dissonance (RQ3)}
\label{sec:rq3_dissonance}

A unique paradox emerges when the model fails. If an adversarial attack successfully fools the model into classifying a fake audio as ``real'', do we want the reasoning to agree with that error? To capture this, we introduce \textbf{Reasoning-Verdict Dissonance} metric to quantify cases where the reasoning layer correctly describes manipulated features $r_i$, but the final verdict $v$ is incorrect.

Let $\mathcal{D}_{\text{Wrong}}$ be the subset of the dataset where the ALM's final prediction $v$ is wrong. We quantify the reasoning-verdict dissonance for a reasoning aspect $r_i$ as the mean disagreement rate between the reasoning trace and the final label within $\mathcal{D}_{\text{Wrong}}$:
\begin{equation}
\label{eq:dissonance}
\Psi_{\text{Diss}}(r_i)
=
\mathbb{E}_{v \sim \mathcal{D}_{\text{Wrong}}}
\left[
1 - \mathcal{E}(r_i, v)
\right],
\end{equation}

\noindent where $\Psi_{\text{Diss}}$ acts as a \textbf{dissonance indicator}: the model made a classification error, but the intermediate reasoning internally flags acoustic anomalies contradicting that label.

\paragraph{Dissonance under Attacks.} By analyzing the dissonance shifts $\Delta \Psi_{\text{Diss}}$, we can understand the model behavior under different adversarial attacks. For instance, if Qwen2-Audio has high $\Delta \Psi_{\text{Diss}}$ under acoustic-based attacks, but low $\Delta \Psi_{\text{Diss}}$ under linguistic-based attacks. Showing that linguistic-based attacks systematically deceive the model without leaving reasoning traces. We compute the metric as:
\begin{equation}
\label{eq:dissonance_shift}
\Delta \Psi_{\text{Diss}}(r_i) = \Psi_{\text{Diss}}^{\text{PER}}(r_i) - \Psi_{\text{Diss}}^{\text{ORG}}(r_i),
\end{equation}

\section{Experimental Set-up}
\noindent \textbf{Datasets.} We evaluate the SARA framework on the ADD ASVspoof 2019 benchmark dataset~\cite{wang2020asvspoof2019largescalepublic}. To adapt this binary classification dataset for reasoning tasks, we are motivated by recent paradigms in synthetic reasoning data generation, such as the distillation protocol of DeepSeek-R1~\cite{Guo2025} and the audio-conditioned QA generation of Audio-Reasoner~\cite{zhifei-etal-2025-audio}. Rather than using complex reinforcement learning pipelines, we employ a simpler, direct process of iterative supervised fine-tuning and generation to synthesize our reasoning training set.

\noindent \textbf{Audio Deepfake Detections.}  For traditional ADD (binary classification), we establish a baseline using AASIST-2, RawNet-2, and CLAD. For ALMs, we prioritize open-source models with robust ecosystem support (HuggingFace and vLLM): Qwen2-Audio-7B-Instruct, Phi-4-multimodal-instruct, gemma-3n-E4B-IT, granite-speech-3.3-8B, and Audio-Flamingo-3-HF. 

\noindent \textbf{Adversarial Attack Frameworks.} To simulate a realistic threat landscape, we employ a dual-perturbation pipeline of linguistic and acoustic attacks. 
For linguistic attacks, we use the TAPAS framework \cite{nguyen-etal-2025-read} across \textbf{four} sub-attacks combining gender (male, female) and accents (American, British). 
Acoustic attacks are evaluated across \textbf{three} protocols defined in CLAD \cite{wu2024cladrobustaudiodeepfake}: Background Noise, Time \& Pitch, and Shape \& Space. 
See Appendix \ref{subsec:attack_def} for detailed parameters and configurations. 

\noindent \textbf{Metrics \& Acronyms.} We report (1) Original Accuracy (OC): detection performance on clean data; and (2) Attack Success Rate (ASR): vulnerability to manipulation. Superscripts $ORG$ and $PER$ denote measurements under clean and perturbed conditions, respectively (e.g., $\Phi_{\text{Coh}}^{PER}$).

Please refer to $\S\ref{subsec:data_synthesis}$, $\S \ref{subsec:data_synthesis}$, and Appendix \ref{subsec:attack_def} for data synthesis, detection, and attack implementation details, respectively.

\section{Experiment Results}

\subsection{Baseline Models}
\label{sec:main_result}

Table \ref{table:main_result} reports ADD performance on the ASVspoof2019 dataset under clean conditions. Traditional binary classifiers establish a strong empirical upper bound, with AASIST-2 achieving $99.58\%$ overall accuracy. When prompted to output direct verdicts without reasoning ($NON$ or \textit{non-reasoning} mode), ALMs match or slightly exceed these baselines. Notably, $\text{gemma-3n-E4B}^{NON}$ and $\text{granite-3.3-8b}^{NON}$ reach ${\sim}99.9\%$ accuracy.

However, chain-of-thought reasoning ($RSN$) degrades overall accuracy by an aggregate of $3.15$\% ($p{=}0.069$) for Granite, Phi-4, and Gemma. This loss stems primarily from degraded bona fide speech detection: F1 score of Fake labels remains stable, while that of Real labels experiences severe declines in Gemma (${-}17.57\%$) and Granite ($-21.0\%$), indicating a strong bias toward false positives in $RSN$ mode. Conversely, Qwen2-Audio remains stable across modes (${\sim} 98.0\%$ accuracy), while Audio-Flamingo-3, one of the recent models reporting state-of-the-art audio reasoning results, serves as a clear outlier, where explicit reasoning substantially lifts accuracy from $92.11\%$ to $97.00\%$. To understand the mechanisms driving these reasoning-performance trade-offs, we begin with Acoustic Perception.

\begin{table}[tb!]
    \footnotesize
    \centering
    \renewcommand{\tabcolsep}{4pt}
    \begin{tabular}{lccc}
        \toprule
        & \textbf{Acc.} & \textbf{Real F1} & \textbf{Fake F1}  \\
        \cmidrule(lr){2-4}

        $\text{Audio-Flamingo-3}^{NON}$ & 92.11\% & 72.35\% & 95.40\% \\
        $\text{Audio-Flamingo-3}^{RSN}$ & 97.00\% & 87.30\% & 98.30\% \\
        \cmidrule(lr){1-4}
        
        $\text{Qwen2-Audio-7B}^{NON}$ & 98.00\% & 91.19\% & 98.88\% \\
        $\text{Qwen2-Audio-7B}^{RSN}$ & \textbf{98.2}\% & \textbf{91.70}\% & \textbf{99.00}\% \\
        
        \cmidrule(lr){1-4}
        $\text{granite-3.3-8b}^{NON}$ & 99.87\% & 99.39\% & 99.93\% \\
        $\text{granite-3.3-8b}^{RSN}$ & 96.11\% & 78.39\% & 97.88\% \\
        
        \cmidrule(lr){1-4}
        $\text{Phi-4-multimodal}^{NON}$ & 97.78\% & 89.42\% & 98.76\% \\
        $\text{Phi-4-multimodal}^{RSN}$ & 96.35\% & 83.01\% & 97.99\% \\
        
        \cmidrule(lr){1-4}
        $\text{gemma-3n-E4B}^{NON}$ & \textbf{99.89}\% & \textbf{99.52}\% & \textbf{99.94}\% \\
        $\text{gemma-3n-E4B}^{RSN}$ & 95.63\% & 81.95\% & 97.73\% \\
    
        \cmidrule(lr){1-4}
        RawNet-2 & 90.86\% & 68.82\% & 94.64\% \\
        CLAD & 98.78\% & 94.37\% & 99.32\% \\
        AASIST-2 & 99.58\% & 98.02\% & 99.77\% \\
        \bottomrule
    \end{tabular}
    \caption{Performance comparison of ALMs on the ASVspoof 2019 task, comparing non-reasoning classification ($NON$) versus reasoning ($RSN$) modes. Best performances of $NON$ and $RSN$ are bold.}
    \label{table:main_result}
    \vspace{-10pt}
\end{table}

\subsection{Acoustic Perception}
\label{subsec:rq1}

\begin{figure}[tb!]
    \centering
    \setlength{\fboxsep}{0pt}
    \includegraphics[width=7.5cm]{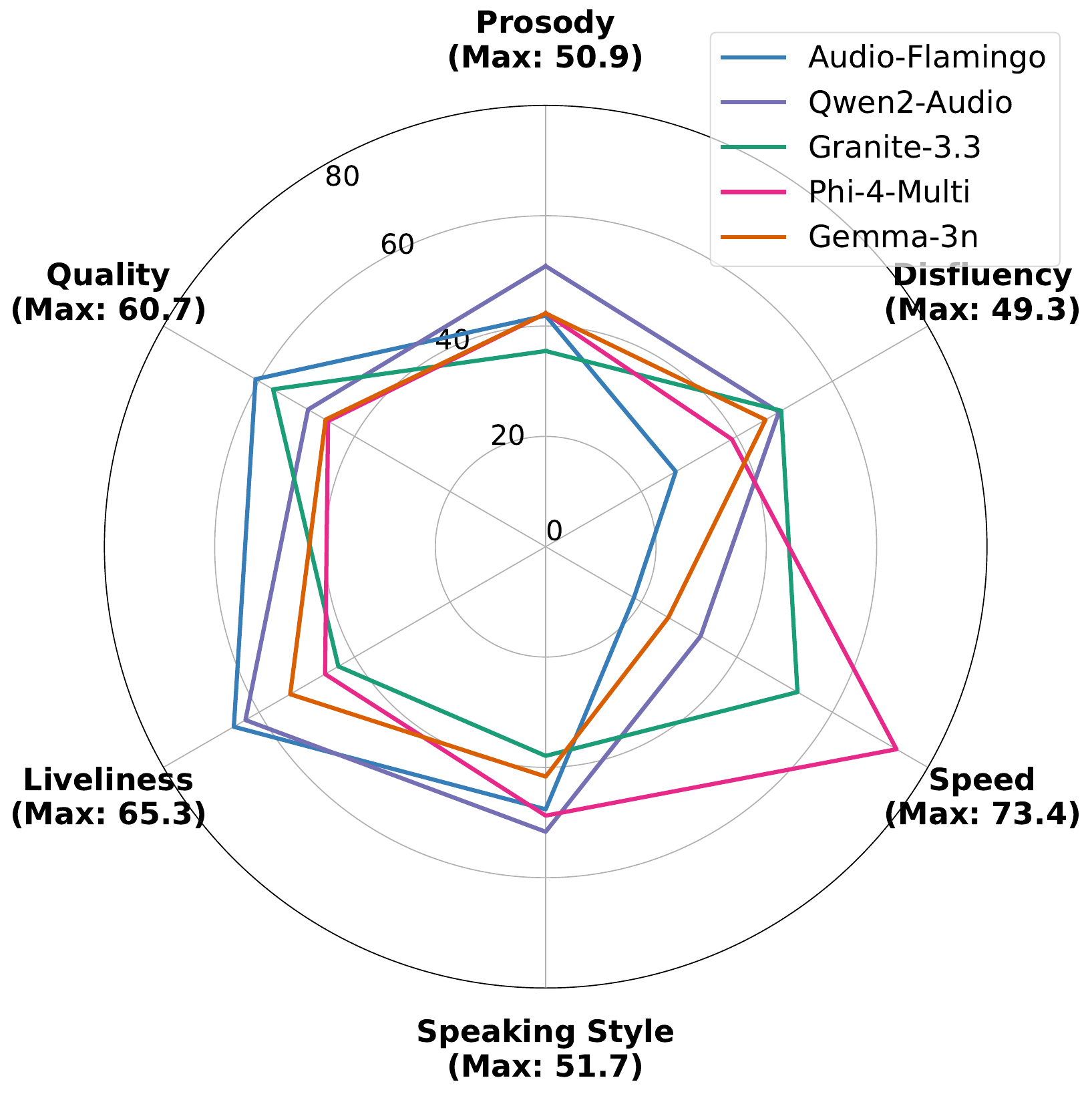}
    \caption{Perception scores ($\Phi_{Perc}$) across six reasoning aspect of five ALMs.}
    \label{fig:rq1_perception}
    \vspace{-10pt}
\end{figure}

Figure \ref{fig:rq1_perception} reports the acoustic perception scores ($\Phi_{\text{Perc}}$) across five ALMs, revealing that foundational acoustic grounding remains a significant challenge. Even the best-performing model, Qwen2-Audio, achieves a mean perception accuracy of only 49.40\%, indicating that current ALMs fail to accurately perceive more than half of the forensic acoustic attributes tested. Averaging across five models, they exhibit the lowest sensitivity to \textbf{Speed} (mean: 40.54\%) while demonstrating their strongest perceptual grounding in \textbf{Liveliness} (mean: 54.24\%). 

Given this limited grounding, we investigate whether perceptual deficits relate to the classification performance trade-offs observed when models generate explicit reasoning in $\S\ref{sec:main_result}$. This analysis is essential to determine whether the accuracy penalties observed in explicit reasoning models are driven by underlying perceptual bottlenecks. We observe a marginal positive correlation between a model's acoustic perception of \textbf{Liveliness} ($\Phi_{\text{Perc}}$) and its classification accuracy in $RSN$ mode ($r {=} 0.827$, $p {=} 0.084$). 
Conversely, acoustic perception of \textbf{Disfluency} exhibits a statistically significant negative correlation with the model's performance when forced to generate reasoning ($r{=}{-}0.945$, $p{=}0.015$), suggesting that models with higher sensitivity to disfluencies are more vulnerable to accuracy degradation in $RSN$ mode.

A natural concern is whether forcing models to generate \textit{explicit reasoning} further distorts these underlying perceptual capabilities. 
Table \ref{table:extended_perception_benchmark} provides a granular comparison of the perception accuracy across the six reasoning dimensions for both the vanilla and finetuned models. 
A \textit{paired t-test} across all perception metrics indicates a \textbf{mean absolute decrease of 1.67\%} for the $RSN$ models. 
However, this degradation is \textbf{not statistically significant} ($t {=} 1.9104$, $p {=} 0.07$). 
This suggests that while explicit reasoning generation introduces a marginal perceptual degradation, it \textit{does not fundamentally compromise} the models' baseline acoustic grounding. 

Ultimately, these findings show that although our synthesis approach enables reasoning capabilities, the models remain bounded by their \textbf{baseline perceptual bottlenecks}. 
This raises a critical question: does explicit reasoning \textit{improve adversarial robustness}, or does it instead \textit{exacerbate the vulnerabilities} of limited acoustic perception?

\subsection{Reasoning Shift under Acoustic Attacks}

\renewcommand{\tabcolsep}{3pt}
\begin{table}[tb!]
    \footnotesize
    \centering
    \begin{tabular}{l c c c c}
    \toprule
    \textbf{ALMs} & \textbf{OC} & \textbf{ASR} & $\Phi_{\text{Coh}}^{PER}$ & $\Psi_{\text{Diss}}^{PER}$ \\

    \cmidrule(lr){2-5}
    Audio-Flamingo-3$^{NON}$ &  94.5 & 25.6 & -- & -- \\
    Audio-Flamingo-3$^{RSN}$ & 97.4 & 46.3 & 85.2 \cohedown{7.8} & 21.3 \dissup{5.3} \\
    
    \cmidrule(lr){1-5}
    Qwen2-Audio-7B$^{NON}$ & 99.1 & 36.6 & -- & -- \\
    Qwen2-Audio-7B$^{RSN}$ & 97.1 & 45.7 & 78.0 \cohedown{8.2} & 29.2 \dissdown{16.8} \\

    \cmidrule(lr){1-5}
    granite-3.3-8b$^{NON}$ & 99.9 & 34.4 & -- & -- \\
    granite-3.3-8b$^{RSN}$ & 82.1 & 49.7 & 73.4 \cohedown{14.6} & 27.8 \dissup{10.6} \\

    \cmidrule(lr){1-5}
    Phi-4-multi$^{NON}$    & 94.4 & 40.7 & -- & -- \\
    Phi-4-multi$^{RSN}$    & 88.8 & 46.1 & 75.5 \cohedown{13.1} & 36.1 \dissup{19.4} \\

    \cmidrule(lr){1-5}
    gemma-3n-E4B$^{NON}$   & 99.8 & 30.6 & -- & -- \\
    gemma-3n-E4B$^{RSN}$   & 77.3 & 49.1 & 43.2 \cohedown{27.4} & 67.9 \dissdown{15.5} \\
    
    \bottomrule
    \end{tabular}
    \caption{Performance under \textbf{Acoustic} Adversarial Attacks. (\cohedown/\coheup) indicate the absolute decrease/increase relative to the original (ORG) performance baseline.}
\label{table:brief_acoustic_attacks}
    \vspace{-10pt}
\end{table}

Table \ref{table:brief_acoustic_attacks} shows that acoustic perturbations induce significant instability in the model's coherency, evidenced by an average degradation in reasoning-verdict coherence ($\Phi_{Coh}$) of 14.20\% ($p {<} 0.0001$). A granular analysis (Table \ref{table:RQ2_coherence}) of the dissonance shift ($\Delta\Psi_{Diss}$) further reveals that these perturbations act as detectable forensic signals within the reasoning traces. Specifically, the dissonance metric, quantifying the conflict between internal reasoning steps and incorrect, final classification, shows marked increases: Audio-Flamingo-3 records an average increase of (\dissup{5.3}) points, Granite-3.3-8b (\dissup{10.6}) points, and Phi-4-multimodal a pronounced (\dissup{19.4}) points. These positive shifts suggest that for many models, the reasoning layer maintains a record of acoustic anomalies even when the classification head is successfully bypassed.

Comparing ASR across settings, $RSN$ mode consistently exhibits higher ASR than the $NON$ baseline, indicating that reasoning-capable ALMs are more susceptible to acoustic manipulation (paired t-test, $p {=} 0.0003$). Finally, to understand the acoustic perception drivers, our correlation analysis (Table \ref{tab:main_combined_corr}, Appendix \ref{app:extended_corr}) reveals that $\Phi_{\text{Perc}}$(\textbf{Speed}) is positively correlated with increased dissonance ($r{=}0.643$, $p{=}0.010$), whereas $\Phi_{\text{Perc}}$(\textbf{Speaking Style}) is positively associated with coherence preservation ($r{=}0.558$, $p{=}0.031$). Although correlation does not imply causation, these patterns suggest that an ALM's capacity to accurately perceive speaking speed and style is crucial for minimizing coherence erosion under attack, informing how we curate training data and fine-tune future models for robust audio reasoning.

\subsection{Reasoning Shift under Linguistic Attacks}

\renewcommand{\tabcolsep}{3pt}
\begin{table}[tb!]
    \footnotesize
    \centering
    \begin{tabular}{lcccc}
\toprule
\textbf{ALMs} & \textbf{OC} & \textbf{ASR} & $\Phi_{\text{Coh}}^{PER}$ & $\Psi_{\text{Diss}}^{PER}$ \\
\cmidrule(lr){2-5}
    Audio-Flamingo-3$^{NON}$ & 35.6 & 95.6 & -- & -- \\
    Audio-Flamingo-3$^{RSN}$ & 81.2 & 90.0 & 73.4 \coheup{3.7} & 25.8 \dissdown{1.6} \\
\cmidrule(lr){1-5}
$\text{Qwen2-Audio-7B}^{NON}$ & 67.4 & 82.8 & -- & -- \\
$\text{Qwen2-Audio-7B}^{RSN}$ & 98.6 & 31.5 & 80.6 \cohedown{7.3} & 9.6 \dissup{6.8} \\
\cmidrule(lr){1-5}
$\text{granite-3.3-8b}^{NON}$ & 35.8 & 94.3 & -- & -- \\
$\text{granite-3.3-8b}^{RSN}$ & 83.6 & 51.8 & 67.0 \cohedown{18.4} & 9.9 \dissdown{12.2} \\
\cmidrule(lr){1-5}
$\text{Phi-4-multimodal}^{NON}$ & 96.6 & 31.1 & -- & -- \\
$\text{Phi-4-multimodal}^{RSN}$ & 91.8 & 52.6 & 69.0 \cohedown{20.5} & 33.4 \dissdown{1.0} \\
\cmidrule(lr){1-5}
$\text{gemma-3n-E4B}^{NON}$ & 72.6 & 54.9 & -- & -- \\
$\text{gemma-3n-E4B}^{RSN}$ & 76.4 & 82.8 & 86.9 \coheup{22.9} & 11.2 \dissdown{1.4} \\
    \bottomrule
    \end{tabular}
    \caption{Performance under \textbf{Linguistic} Adversarial Attacks.}
    \label{table:brief_linguistic_attacks}
    \vspace{-10pt}
\end{table}

\begin{table}[tb!]
\footnotesize
\centering
\begin{tabular}{lcccc}
\toprule
& \multicolumn{2}{c}{\textbf{Acoustic Attacks}} 
& \multicolumn{2}{c}{\textbf{Linguistic Attacks}} \\
\cmidrule(lr){2-3} \cmidrule(lr){4-5}
\textbf{Dimension} 
& $r_{\text{Coh}}$ & $r_{\text{Diss}}$
& $r_{\text{Coh}}$ & $r_{\text{Diss}}$ \\
\midrule
Prosody        
& 0.274 
& \sigA{-0.566}  
& 0.156     
& \sigB{0.602} \\

Disfluency     
& -0.353 
& -0.362  
& -0.115    
& -0.052  \\

Speed          
& 0.049 
& \sigB{0.643}  
& \sigC{-0.727} 
& -0.187  \\

Speaking Style 
& \sigA{0.558} 
& -0.127  
& -0.121    
& \sigA{0.541}  \\

Liveliness     
& 0.381 
& -0.502  
& 0.430     
& 0.428   \\

Quality        
& 0.475 
& 0.220   
& -0.097    
& -0.262  \\
\bottomrule
\end{tabular}
\caption{Pearson correlation analysis of $\Phi_{\text{Perc}}$ against reasoning coherence and dissonance shifts under acoustic and linguistic adversarial attacks. Cell background intensity indicates statistical significance: lighter to darker shading corresponds to $p{<}0.05$, $p{<}0.01$, and $p{<}0.001$, respectively.}
\label{tab:main_combined_corr}
\vspace{-15pt}
\end{table}

Under linguistic perturbations (Table \ref{table:brief_linguistic_attacks}), explicit reasoning marginally reduces the aggregate ASR from $71.75\%$ to $61.73\%$ ($p{=}0.27$). This shift shows negligible changes in reasoning-verdict coherence ($\Delta\Phi_{\text{Coh}} {=}{-}3.91$, $p{=}0.33$) and dissonance ($\Delta\Psi_{\text{Diss}} {=}{-}4.03$, $p{=}0.143$). Thus, linguistic attacks \textit{do not trigger the same reasoning friction} observed under acoustic threats, instead slightly decreasing internal dissonance. Compared to acoustic attacks, linguistic variants yield a higher aggregate ASR ($61.73\%$ vs. $47.38\%$) while inducing lower coherence erosion and minimal dissonance signals.

Comparing ASR across $NON$ and $RSN$ modes reveals that while $RSN$ exhibits higher mean ASR than the $NON$ baseline, this difference is not statistically significant ($p {=} 0.27$). This suggests the impact of reasoning on linguistic attack vulnerability is less systemic than in the acoustic regime. Finally, extended analysis in Table \ref{tab:main_combined_corr} demonstrates that higher $\Phi_{\text{Perc}}$(\textbf{Speed}) is negatively associated with coherence stability ($r{=}{-}0.727$, $p{<}0.001$), while $\Phi_{\text{Perc}}$(\textbf{Prosody}) is positively correlated with reasoning-verdict dissonance ($r{=}0.602$, $p{=}0.005$). These correlations suggest that linguistic perturbations, unlike acoustic attack strategies, systematically exploit specific perceptual-semantic misalignment in the model’s internal logic. Consequently, while acoustic attacks trigger detectable logical friction, linguistic attacks silently deceive both the reasoning chain and the final verdict, indicating that future ALM training must prioritize robust cross-modal alignment rather than relying solely on textual coherence checks.

\renewcommand{\tabcolsep}{7pt}
\begin{table}[tb!]
    \footnotesize
    \centering
\begin{tabular}{lccc}
\toprule
 & \textbf{Coef} & \textbf{dy/dx} & \textbf{P>|z|} \\
\cmidrule(lr){2-4}

Prediction Score   
& 0.5391 
& 0.1229  
& 0.000 \\

\cmidrule(lr){1-4}

$\Phi_{\text{Coh}}$(Prosody)  
& \topneg{-0.2660} 
& -0.0607 
& 0.000 \\

$\Phi_{\text{Coh}}$(Disfluency)  
& -0.1452 
& -0.0331 
& 0.000 \\

$\Phi_{\text{Coh}}$(Speed)   
& 0.0291  
& 0.0066  
& 0.004 \\

$\Phi_{\text{Coh}}$(Speaking Style) 
& \toppos{0.0447}
& 0.0102 
& 0.000 \\

$\Phi_{\text{Coh}}$(Liveliness)   
& -0.1351 
& -0.0308 
& 0.000 \\

$\Phi_{\text{Coh}}$(Quality)  
& 0.0191  
& 0.0044 
& 0.041 \\

\cmidrule(lr){1-4}

$\Phi_{\text{Perc}}$(Liveliness)    
& \topneg{-0.2105}
& -0.0480 
& 0.000 \\

$\Phi_{\text{Perc}}$(Quality)       
& 0.0895  
& 0.0204 
& 0.000 \\

$\Phi_{\text{Perc}}$(Speaking Style) 
& \toppos{0.2115}  
& 0.0482  
& 0.000 \\

\bottomrule
\end{tabular}
\caption{Logistic regression for adversarial attack detection across 87,560 samples (35 ALM-attack pairs). Collinear features were removed via stepwise VIF elimination (threshold = 10.0). \textit{Prediction Score} is model confidence, computed by normalizing the generation log-probabilities of the verdict token ($fake$ and $real$).}
\label{tab:adv_atk_prediction}
\vspace{-10pt}
\end{table}


\section{Discussion}

\subsection{Reasoning Coherence for Adversarial Attack Detection}
\label{subsec:attack_detection}

Beyond coherence-dissonance diagnosis, the SARA framework also suggests a potential application in adversarial defense: \textit{``Can reasoning traces serve as standalone signals for detecting adversarial perturbations, even when the ALM's final prediction is deceived?''} To evaluate this, we train a binary classifier to identify adversarial inputs using prediction scores, acoustic perceptions, and reasoning-coherence metrics as predictive features.

Regression results show that adversarial perturbations are associated with degraded reasoning coherence (Table~\ref{tab:adv_atk_prediction}). Several acoustic dimensions have significant negative coefficients, especially prosody ($\beta{=}-0.266$, $p{<}0.001$), disfluency ($\beta{=}-0.145$, $p{<}0.001$), and liveliness ($\beta{=}-0.135$, $p{<}0.001$). Lower coherence in these dimensions increases the predicted likelihood of adversarial input, with prosody degradation corresponding to a 6.07\% increase. This suggests that attacks expose dimension-specific reasoning vulnerabilities, which may also affect related speech tasks such as emotion recognition.

Table~\ref{tab:adv_atk_prediction_performance} confirms this approach, with LightGBM achieving a 77.59\% aggregate F1 score. Performance varies by attack modality: linguistic attacks produce less overall coherence erosion than acoustic perturbations, yet are detected more accurately (87.10\% vs. 76.97\% F1; Tables~\ref{tab:linguistic_regression} and~\ref{tab:acoustic_regression}). This suggests that linguistic attacks, though stealthier to standard classifiers, leave a distinct signature in reasoning traces, demonstrating SARA's value beyond diagnosis as a practical tool for adversarial detection.



\renewcommand{\tabcolsep}{5pt}
\begin{table}[tb!]
    \footnotesize
    \centering
\begin{tabular}{lccc}
\toprule
\textbf{Model} & \textbf{Precision} & \textbf{Recall} & \textbf{F1 Score} \\ \cmidrule(lr){2-4}
LightGBM             & 76.72\% & 78.48\% & \textbf{77.59}\% \\
XGBoost              & 76.83\% & 76.77\% & 76.80\% \\
Random Forest        & 75.48\% & 72.35\% & 73.88\% \\
AdaBoost             & 69.53\% & 68.25\% & 68.89\% \\ 
SVM    & 70.45\% & 59.42\% & 64.46\% \\
Logistic Regression  & 65.24\% & 60.19\% & 62.61\% \\

\bottomrule
\end{tabular}
\caption{Performance of adversarial attack detection based on reasoning traces.}
\label{tab:adv_atk_prediction_performance}
\vspace{-15pt}
\end{table}

\subsection{Sufficiency of Reasoning Data Synthesis}
\label{subsec:data_synthesis}


Reasoning in ALMs remains an early and underdeveloped area, with current systems still showing a substantial gap compared with reasoning capabilities in vision-language and text-only language models. This makes obtaining reliable reasoning supervision is especially challenging. We acknowledge that our use of AI-synthesized CoT traces is a constraint of this study. Human-expert annotations would provide stronger diagnostic reliability, but they are costly to produce at scale and were beyond the primary scope of this work, which aims to propose a general diagnostic framework for ALMs' reasoning. 

\begin{figure}[t]
    \centering
    \includegraphics[width=7.5cm]{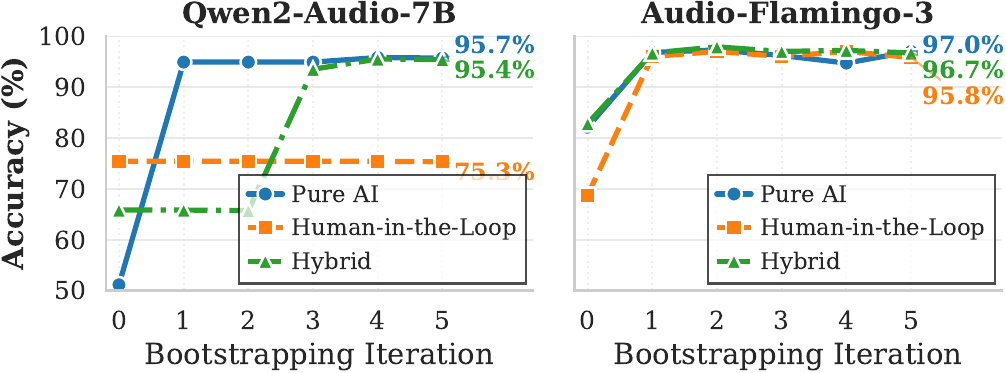}
    \caption{Seed Data Generation Comparison by Qwen-2-Audio-7B and Audio-Flamingo-3}
\label{fig:seed_reasoning_data_synthetic_comparison}
\vspace{-15pt}
\end{figure}

Empirically, models adapted more effectively to human-taxonomy-guided synthetic traces than to the reasoning patterns in human-provided labels. Evidently, Figure \ref{fig:seed_reasoning_data_synthetic_comparison} illustrates the performance of Qwen-2-Audio-7B and Audio-Flamingo-3 over five iterative refinement steps across three seed variants: Human-labeled (100 samples), Pure AI (100 samples), and Hybrid (mixing both; 200 samples). We observed that the Pure AI seed, which is curated from human annotation insights~\cite{10.1145/3658644.3670325} following protocol similar to DeepSeek-R1~\cite{Guo2025}, that we utilized in this paper consistently outperformed human-manual-labeled samples, making it the best approach for synthesizing reasoning for ALM fine-tuning. 

Although synthetic reasoning generation is not the central contribution of this work, we present it as an enabling step for this diagnostic study and encourage future work to validate reasoning-based defenses with larger, human-curated audio reasoning datasets.

\subsection{Generalizability Across Synthetic Data Sources}

To ensure our findings are not artifacts of our primary synthetic dataset (generated via Qwen2-Audio-7B), we evaluate SARA on an independent transfer dataset synthesized using Audio-Flamingo-3 (Table \ref{table:extended_result_af3}). The results confirm that the observed reasoning patterns are dataset-agnostic. For example, the classification penalties under RSN mode, such as Qwen2-Audio dropping to $95.9\%$ and Granite to $98.9\%$ (Table \ref{table:extended_result_af3}), closely mirror the trends on our primary set. Crucially, the divergence between attack modalities persists: acoustic attacks consistently degrade internal coherency, while linguistic attacks preserve coherence across both synthetic sources.


However, the adversarial detector ($\S$\ref{subsec:attack_detection}) exhibits sensitivity to the synthetic generator, yielding a low zero-shot F1 score of $53.65\%$ when directly evaluated on the AF3-synthesized results. In practice, this can be mitigated by aligning training data with the target model's output distribution to ensure deployment robustness.

\section{Conclusion}
We introduced SARA to analyze ALM reasoning to detect deepfake under adversarial attacks. Our study reveals that reasoning traces can capture internal logic shifts that occur during an attack. When an adversary successfully deceives a model's final verdict, the reasoning often exposes the manipulation through internal contradictions. By measuring this dissonance, SARA acts as a fail-safe to surface perturbed inputs evading standard classifiers. This underscores that inspecting intermediate logic is essential for robust, transparent deepfake detection.

\clearpage
\section*{Limitations}
Our experiments are exclusively conducted on English-language datasets, it remains an open question how our framework generalizes to diverse syntactic and morphological structures of non-English languages.

Due to computational constraints, this study focuses on mid-sized Audio Language Models (approx. 7B-8B parameters) and a small subset of traditional ADD. We do not examine the behavior of large omni-models, such as Qwen3-Omni-30B-A3B-Instruct, nor do we benchmark against an exhaustive list of legacy classifiers. 

Finally, our framework relies on AI-synthesized Chain-of-Thought (CoT) traces, which bounds the generalizability of our findings to human-annotated reasoning paradigms. While human-expert annotations remain the gold standard, they are prohibitively expensive to produce at scale and were outside the primary scope of this diagnostic study. We explored hybrid datasets integrating human-labeled seeds, yet observed that these did not yield performance gains over pure synthetic generation. This suggests a potential alignment bias, where models may adapt more effectively to the consistent logical structures of synthetic traces than to the correctness of human-provided reasoning. We view this not as a failure of our approach, but as a known constraint of using current reasoning-capable ALMs as both the student and the teacher.

\section*{Ethical Considerations}

By shifting the paradigm from ``black-box'' classification to ``glass-box'' reasoning, our work aims to restore trust in automated deepfake detection. The introduction of interpretable metrics like \textit{reasoning-verdict dissonance} empowers human analysts to verify AI decisions rather than blindly accepting them. This auditability is crucial for deploying ALMs in high-stakes environments where a binary fake/real label is insufficient.

We acknowledge that detailing specific \textit{Linguistic Attacks} (e.g., transcript-based perturbations) presents a dual-use risk. Malicious actors could leverage our findings to craft stealthy deepfakes that bypass reasoning-based defenses by exploiting the semantic-acoustic gap. However, we believe that defensive disclosure is necessary. By exposing the fragility of the current ALMs, we enable the community to develop robust alignment techniques before these vulnerabilities are exploited in the wild.

This research is intended solely to strengthen the defense of digital media integrity. We are releasing our attack protocols and audit code to facilitate reproducibility and accelerate the development of red-teaming benchmarks for Audio Language Models. We strongly condemn the use of these techniques for deception or manipulation.

\clearpage

\bibliography{custom}

\clearpage
\appendix

\section{Appendix}
\label{sec:appendix}

\subsection{Attack Strategy Definition}
\label{subsec:attack_def}

To rigorously stress-test Audio Language Models (ALMs), we employ a dual attack framework consisting of Linguistic (Text-based) and Acoustic (Signal-based) perturbations.

\paragraph{Linguistic Adversarial Attacks (TAPAS).}
We utilize the TAPAS framework to generate adversarial audio via a text-to-speech pipeline. 
\begin{itemize}
    \item \textbf{Perturbation Method:} We employ \textbf{TextFooler}, a black-box attack that replaces words with synonyms to maximize semantic similarity while minimizing classification accuracy. We select TextFooler over other candidates (e.g., PWWS, BERTAttack) due to its high empirical success rate in our preliminary screenings.
    \item \textbf{TTS System:} We utilize \textbf{Kokoro TTS} for audio synthesizer. This choice is driven by computational efficiency essential for large-scale reasoning generations; Kokoro is capable of generating 10,000 audio samples in approximately 832 seconds on a single NVIDIA A100 GPU.
    \item \textbf{Attack Example:} \textit{Original:} ``She spoke clearly.'' $\rightarrow$ \textit{Adversarial Transcript:} ``She spoke \textit{flawlessly}.'' $\rightarrow$ \textit{Result:} The acoustic output retains an original voice but contains complex prosody derived from the verbose text.
\end{itemize}

\paragraph{Acoustic Adversarial Attacks.}
We adopt the acoustic perturbation protocols defined in CLAD, organized into three specific recipes used in our \texttt{AcousticAttacker} module:

\noindent \textbf{1. Background Noise Recipe ($\mathcal{A}_{\text{noise}}$)} \\
This strategy tests the model's ability to separate speech from interference.
\begin{itemize}
    \item \textbf{White Noise:} Gaussian noise added to the signal. Signal-to-Noise Ratio (SNR) is sampled uniformly $U \sim [15, 25]$ dB.
    \item \textbf{Environmental Noise:} Real-world background audio (wind, footsteps, breathing, coughing, rain, clock ticks, sneezing) mixed with the source. SNR is sampled uniformly $U \sim [5, 20]$ dB.
\end{itemize}

\noindent \textbf{2. Time \& Pitch Recipe ($\mathcal{A}_{\text{time}}$)} \\
This strategy targets temporal alignment and frequency perception.
\begin{itemize}
    \item \textbf{Time Stretch:} The waveform is stretched or compressed without altering pitch using Phase Vocoding. Ratios are sampled from $\{0.90\times, 0.95\times, 1.05\times, 1.10\times\}$.
    \item \textbf{Time Shift:} The audio is cyclically rolled along the time axis. Shift magnitudes are sampled from $\{1600, 16000, 32000\}$ samples (corresponding to 0.1s, 1s, and 2s at 16kHz).
\end{itemize}

\noindent \textbf{3. Shape \& Space Recipe ($\mathcal{A}_{\text{shape}}$)} \\
This strategy manipulates the signal envelope and spatial characteristics.
\begin{itemize}
    \item \textbf{Volume Change:} Amplitude scaling factor sampled uniformly $U \sim [0.5, 2.0]$.
    \item \textbf{Fade Effects:} Application of linear, logarithmic, exponential, or sine-based fade-in/fade-out envelopes. Fade duration is fixed up to 50\% of the audio length.
    \item \textbf{Synthetic Echo:} A delayed superposition of the signal $x(t) \leftarrow x(t) + \alpha \cdot x(t-\delta)$. Delay $\delta$ is sampled between $[1000, 2000]$ samples; strength $\alpha \sim [0.2, 0.5]$.
\end{itemize}

\subsection{Implementation Details}
\label{subsec:implementation}
\begin{table*}[tb!]
    \footnotesize
    \centering
    \renewcommand{\tabcolsep}{3pt}
    \begin{tabular}{lccccccccc}
        \toprule
        & \textbf{Acc.} & 
        $\textbf{OC}^{acu}$ & 
        $\textbf{ASR}^{acu}$ & $\Phi_{\text{Coh}}^{PER}$ & $\Psi_{\text{Diss}}^{PER}$ & $\textbf{OC}^{lng}$ & 
        $\textbf{ASR}^{lng}$ & $\Phi_{\text{Coh}}^{PER}$ & $\Psi_{\text{Diss}}^{PER}$\\
        \cmidrule(lr){2-2} \cmidrule(lr){3-6} \cmidrule(lr){7-10}
        $\text{Qwen2-Audio-7B}^{RSN}$ & 95.9 & 97.4 & 42.6 & 87.5 \cohedown{-5.7} & 14.8 \dissup{5.8} & 99.4 & 3.9 & 88.8 \cohedown{-0.7} & 38.0 \dissup{25.1} \\
        $\text{granite-3.3-8b}^{RSN}$ & 98.9 & 92.8 & 48.5 & 80.4 \cohedown{-10.0} & 21.8 \dissup{5.3} & 99.5 & 16.3 & 88.0 \cohedown{-2.7} & 18.7 \dissdown{-28.9} \\
        $\text{Phi-4-multimodal}^{RSN}$ & 97.1 & 92.8 & 53.3 & 86.1 \cohedown{-5.6} & 17.6 \dissup{1.4} & 98.9 & 11.7 & 88.9 \coheup{3.0} & 3.4 \dissup{1.3} \\
        $\text{gemma-3n-E4B}^{RSN}$ & 52.9 & 50.8 & 85.7 & 51.4 \coheup{27.9} & 48.0 \dissdown{-34.8} & 92.8 & 93.1 & 89.5 \coheup{7.1} & 5.5 \dissup{0.5} \\
        \bottomrule
    \end{tabular}
    \caption{Performance comparison of Audio Language Models (ALMs) on the ASVspoof 2019 deepfake detection task synthesized by Audio-Flamingo-3.}
    \label{table:extended_result_af3}
    \vspace{-10pt}
\end{table*}

\paragraph{Model Fine-tuning and QLoRA.}
All Audio Language Models (ALMs) were fine-tuned using the HuggingFace \texttt{transformers} library on a single NVIDIA A100 GPU. We employ an instruction-tuning approach where the loss is calculated exclusively on the \textit{Assistant} tokens (the reasoning and verdict), while masking the \textit{System} prompt and \textit{User} audio inputs. To manage the memory constraints of reasoning-heavy generation, we employed \textbf{QLoRA} (Quantized Low-Rank Adaptation). 

Key hyperparameters include:
\begin{enumerate}[noitemsep, nolistsep]
    \item \textbf{Precision:} bfloat16 (BF16).
    \item \textbf{Optimization:} AdamW optimizer with $\beta_1=0.9, \beta_2=0.95$. Learning rate is set to $1\text{e-}4$ with a linear decay scheduler.
    \item \textbf{Batch Size:} Global batch size of 16 (2 per device $\times$ gradient accumulation).
    \item \textbf{Regularization:} Weight decay of $0.1$.
    \item \textbf{Training Duration:} Models are trained until convergence (variable epochs) without early stopping.
\end{enumerate}

The Low-Rank Adaptation (LoRA) hyperparameters were configured as follows:
\begin{enumerate}[noitemsep, nolistsep]
    \item \textbf{Rank ($r$):} 16
    \item \textbf{Alpha ($\alpha$):} 64
    \item \textbf{Dropout:} 0.05
    \item \textbf{Target Modules:} Query, Key, Value, and Output projection layers (\texttt{[q, k, v, o]\_proj})
\end{enumerate}

\subsection{Implementation Details of The Entailment Function}
\label{app:imp_entailment}
As defined in Section 3.3, the entailment function $\mathcal{E}(r_i, v) \rightarrow \{1, 0\}$ is designed to automatically verify whether the intermediate reasoning trace for a specific aspect $r_i$ entails the model's final classification verdict $v \in \{\text{fake}, \text{real}\}$. 

To implement this reliably at scale, we formalize the evaluation as a multi-label classification task. We employ \textbf{Qwen3-8B} as the backbone Large Language Model (LLM) and fine-tune it to systematically analyze the semantic stance of the generated reasoning traces across all six forensic dimensions defined in our taxonomy (Prosody, Disfluency, Speed, Speaking Style, Liveliness, and Quality). 

Specifically, the model takes the entire reasoning trace as input and predicts a binary directional label $p_i \in \{0, 1\}$ for each aspect $i \in \{1, \dots, 6\}$. The label mapping is defined as follows:
\begin{itemize}
    \item $p_i = 0$: The reasoning text $r_i$ articulates evidence that the audio is \textit{fake}..
    \item $p_i = 1$: The reasoning text $r_i$ articulates evidence that the audio is \textit{real}.
\end{itemize}

Once the directional label $p_i$ is extracted for a given aspect, the entailment function compares this semantic stance against the ALM's final predicted verdict $v$ (where we map $v_{\text{fake}} = 0$ and $v_{\text{real}} = 1$). The logical relation is formalized as a simple equivalence check:
\begin{equation}
    \mathcal{E}(r_i, v) = \mathbb{I}(p_i = v)
\end{equation}
where $\mathbb{I}(\cdot)$ is the indicator function. If the premise aligns with the conclusion ($p_i = v$), the aspect is deemed coherent ($\mathcal{E} = 1$). If the reasoning contradicts the final decision (e.g., reasoning describes synthetic artifacts but predicts ``Real''), it is flagged as a dissonance ($\mathcal{E} = 0$). To ensure the automated entailment evaluator's rigor and reliability, we validated the fine-tuned Qwen3-8B model on a held-out set of 100 human-annotated reasoning traces. On this validation set, the classifier achieved a highly robust \textbf{micro-averaged F1 score (\texttt{f1-micro}) of 95.19\%}.

\subsection{Extended Experimental Results}

\begin{table*}[tb!]
    \footnotesize
    \centering
    \renewcommand{\tabcolsep}{3pt}
    \begin{tabular}{lcccccc}
        \toprule
        & \textbf{Disfluency} & \textbf{Liveliness} & \textbf{Prosody} & \textbf{Quality} & \textbf{Speaking Style} & \textbf{Speed} \\
        \cmidrule(lr){2-7}
        $\text{Phi-4-multimodal}$$^{VANILLA}$ & 38.97\% & 46.18\% & 42.30\% & 45.55\% & 48.75\% & 73.42\% \\
        $\text{Phi-4-multimodal}$$^{TRAINED}$ & 40.38\% & 44.79\% & 40.96\% & 46.07\% & 48.75\% & 69.82\% \\
        \cmidrule(lr){1-7}
        $\text{Qwen2-Audio-7B}$$^{VANILLA}$ & 48.83\% & 62.85\% & 50.87\% & 49.74\% & 51.67\% & 32.43\% \\
        $\text{Qwen2-Audio-7B}$$^{TRAINED}$ & 51.17\% & 64.24\% & 43.91\% & 51.31\% & 50.52\% & 34.68\% \\
        \cmidrule(lr){1-7}
        $\text{gemma-3n-E4B}$$^{VANILLA}$ & 46.01\% & 53.47\% & 42.30\% & 46.07\% & 41.67\% & 25.68\% \\
        $\text{gemma-3n-E4B}$$^{TRAINED}$ & 42.72\% & 45.83\% & 36.55\% & 38.74\% & 36.15\% & 27.03\% \\
        \cmidrule(lr){1-7}
        $\text{granite-3.3-8b}$$^{VANILLA}$ & 49.30\% & 43.40\% & 35.48\% & 57.07\% & 37.92\% & 52.70\% \\
        $\text{granite-3.3-8b}$$^{TRAINED}$ & 56.81\% & 46.18\% & 35.07\% & 56.02\% & 37.71\% & 38.29\% \\
        \cmidrule(lr){1-7}
        $\text{Audio-Flamingo-3}$$^{VANILLA}$ & 27.23\% & 65.28\% & 41.90\% & 60.73\% & 47.60\% & 18.47\% \\
        $\text{Audio-Flamingo-3}$$^{TRAINED}$ & 22.07\% & 62.15\% & 38.55\% & 50.79\% & 52.19\% & 24.32\% \\
    \bottomrule
    \end{tabular}
    \caption{Performance of acoustic perception benchmark on VANILLA and finetuned Audio Language Models (ALMs).}
\label{table:extended_perception_benchmark}
\end{table*}

\subsubsection{Acoustic Adversarial Attacks Results}
\renewcommand{\tabcolsep}{5pt}
\begin{table*}[tb!]
    \footnotesize
    \centering
    \begin{tabular}{lccccccccc}
        \toprule
        \textbf{ALMs} & \textbf{Strategy} & \multicolumn{2}{c}{\textbf{NON}} & \multicolumn{6}{c}{\textbf{RSN}} \\
        & & \textbf{OC}$\uparrow$ & \textbf{ASR}$\downarrow$ & \textbf{OC}$\uparrow$ & \textbf{ASR}$\downarrow$ & $\Phi_{\text{Coh}}^{ORG}$ &
        $\Phi_{\text{Coh}}^{PER}$ &
        $\Psi_{\text{Diss}}^{ORG}$ &
        $\Psi_{\text{Diss}}^{PER}$ \\
        
        \cmidrule(lr){2-2} \cmidrule(lr){3-4} \cmidrule(lr){5-10}
        $\text{Audio-Flamingo-3}$ & Background Noise & 94.5 & 28.2 & 97.4 & 56.4 &  93.0 & 85.3 & 16.0 & 16.6\\
        $\text{Audio-Flamingo-3}$ & Shape Space & 94.5 & 29.5 & 97.4 & 47.3 & 93.0 & 85.4 & 16.0 & 20.0\\
        $\text{Audio-Flamingo-3}$ & Time Pitch & 94.5 & 19.2 & 97.4 & 35.1 & 93.0 & 85.0 & 16.0 & 27.4\\
        \cmidrule(lr){1-10}
        
        $\text{Qwen2-Audio-7B}$ & Background Noise & 99.1 & 50.3 & 97.1 & 54.2 &  86.2 & 77.6 & 46.0 & 27.3\\
        $\text{Qwen2-Audio-7B}$ & Shape Space & 99.1 & 43.5 & 97.1 & 49.3 &  86.2 & 80.5 & 46.0 & 25.0\\
        $\text{Qwen2-Audio-7B}$ & Time Pitch & 99.1 & 16.0 & 97.1 & 33.7 &  86.2 & 76.0 & 46.0 & 35.4\\
        \cmidrule(lr){1-10}
        
        $\text{Phi-4-multimodal}$ & Background Noise & 94.4 & 34.5 & 88.8 & 44.8 &  88.4 & 72.1 & 15.9 & 41.3\\
        $\text{Phi-4-multimodal}$ & Shape Space & 94.4 & 39.9 & 88.8 & 46.8 & 88.5 & 74.2 & 17.9 & 40.1\\
        $\text{Phi-4-multimodal}$ & Time Pitch & 94.4 & 47.8 & 88.8 & 46.8 & 88.9 & 80.3 & 16.2 & 26.9 \\
        \cmidrule(lr){1-10}
        
        $\text{gemma-3n-E4B}$ & Background Noise & 99.8 & 41.7 & 77.3 & 51.1 &  72.0 & 49.9 & 82.6 & 55.6\\
        $\text{gemma-3n-E4B}$ & Shape Space & 99.8 & 16.0 & 77.3 & 48.1 & 69.1 & 37.6 & 85.8 & 78.2\\
        $\text{gemma-3n-E4B}$ & Time Pitch & 99.8 & 34.1 & 77.3 & 48.1 & 70.8 & 42.2 & 81.8 & 69.8\\
        \cmidrule(lr){1-10}
        
        $\text{granite-3.3-8b}$ & Background Noise & 99.9 & 12.1 & 82.1 & 48.6 & 87.6 & 73.0 & 15.3 & 28.2\\
        $\text{granite-3.3-8b}$ & Shape Space & 99.9 & 50.0 & 82.1 & 50.0 & 88.3 & 68.8 & 17.8 & 36.7\\
        $\text{granite-3.3-8b}$ & Time Pitch & 99.9 & 41.2 & 82.1 & 50.5 & 88.0 & 78.3 & 18.5 & 18.6\\
        \bottomrule
    \end{tabular}
    \caption{Performances under \textbf{Acoustic Adversarial Attacks}.}
    \label{table:acoustic_attacks}
    \vspace{-10pt}
\end{table*}

Table \ref{table:acoustic_attacks} demonstrates that signal-level perturbations systematically destabilize the reasoning chain. We observe substantial degradation of coherence across multiple models; for example, gemma-3n-E4B exhibits a significant decline in logical consistency under Shape Space attacks (from $\Phi_{\text{Coh}}^{\text{ORG}} =69.1\%$ to $\Phi_{\text{Coh}}^{\text{PER}} =37.6\%$), suggesting a failure to maintain a stable logical chain when processing perturbed inputs. Conversely, Phi-4-multimodal provides diagnostic utility through increased dissonance; under Background Noise attacks, despite an Attack Success Rate of 44.8\%, the model exhibits a pronounced increase in dissonance ($\Psi_{\text{Diss}}^{\text{PER}} = 41.3\%$), indicating internal logical conflict following a classification failure. Qwen2-Audio maintains the highest robustness, sustaining coherence between 77-80\% and demonstrating minimal dissonance shifts across all evaluated acoustic attack vectors.

\subsubsection{Linguistic Adversarial Attacks Results}
\renewcommand{\tabcolsep}{5pt}
\begin{table*}[tb!]
    \footnotesize
    \centering
    \begin{tabular}{lccccccccc}
        \toprule
        \textbf{ALMs} & \textbf{Voice Profile} & \multicolumn{2}{c}{\textbf{NON}} & \multicolumn{6}{c}{\textbf{RSN}} \\
        & & \textbf{OC}$\uparrow$ & \textbf{ASR}$\downarrow$ & \textbf{OC}$\uparrow$ & \textbf{ASR}$\downarrow$ & $\Phi_{\text{Coh}}^{ORG}$ &
        $\Phi_{\text{Coh}}^{PER}$ &
        $\Psi_{\text{Diss}}^{ORG}$ &
        $\Psi_{\text{Diss}}^{PER}$ \\
        
        \cmidrule(lr){2-2} \cmidrule(lr){3-4} \cmidrule(lr){5-10}
        $\text{Audio-Flamingo-3}$ & American Female & 0 & 100 & 65.1 & 98.1 &  65.9 & 74.4 & 24.2 & 25.4\\
        $\text{Audio-Flamingo-3}$ & American Male & 46.9 & 97.6 & 92.1 & 81.4 & 71.8 & 72.8 & 31.4 & 25.5\\
        $\text{Audio-Flamingo-3}$ & British Female & 4.0 & 100 & 77.0 & 96.0 & 70.8 & 73.9 & 25.4 & 25.8\\
        $\text{Audio-Flamingo-3}$ & British Male & 91.3 & 84.8 & 90.6 & 84.5 & 70.4 & 72.6 & 28.4 & 26.4\\
        \cmidrule(lr){1-10}
        
        $\text{Qwen2-Audio-7B}$ & American Female & 44.9 & 98.9 & 98.3 & 26.3 &  87.4 & 80.2 & 10.3 & 9.6\\
        $\text{Qwen2-Audio-7B}$ & American Male & 98.9 & 52.1 & 98.9 & 52.1 & 84.8 & 77.2 & 0.0 & 9.3\\
        $\text{Qwen2-Audio-7B}$ & British Female & 39.2 & 98.4 & 98.4 & 28.1 & 95.7 & 87.5 & 0.0 & 12.9\\
        $\text{Qwen2-Audio-7B}$ & British Male & 86.8 & 81.8 & 98.8 & 19.5 & 83.6 & 77.3 & 0.9 & 6.5\\
        \cmidrule(lr){1-10}
        
        $\text{Phi-4-multimodal}$ & American Female & 88.3 & 69.2 & 72.8 & 94.7 & 82.5 & 64.3 & 34.1 & 36.2\\
        $\text{Phi-4-multimodal}$ & American Male & 98.1 & 31.8 & 95.5 & 55.8 & 91.4 & 72.0 & 26.2 & 28.0\\
        $\text{Phi-4-multimodal}$ & British Female & 100.0 & 5.6 & 99.4 & 39.5 & 91.3 & 67.4 & 20.4 & 38.5\\
        $\text{Phi-4-multimodal}$ & British Male & 99.9 & 17.9 & 99.5 & 20.5 & 92.7 & 72.1 & 57.1 & 30.9\\
        \cmidrule(lr){1-10}
        
        $\text{gemma-3n-E4B}$ & American Female & 3.1 & 100 & 54.4 & 100.0 & 50.0 & 95.3 & 4.3 & 4.7\\
        $\text{gemma-3n-E4B}$ & American Male & 93.1 & 55.4 & 86.0 & 74.9 & 71.5 & 82.1 & 14.6 & 14.4\\
        $\text{gemma-3n-E4B}$ & British Female & 94.7 & 62.3 & 75.8 & 89.1  & 61.6 & 87.8 & 11.3 & 10.7\\
        $\text{gemma-3n-E4B}$ & British Male & 99.4 & 2.0 & 89.4 & 67.0 & 72.7 & 82.5 & 20.0 & 15.2\\
        \cmidrule(lr){1-10}

        $\text{granite-3.3-8b}$ & American Female & 0.5 & 100 & 97.4 & 40.7 & 84.9 & 67.9 & 16.2 & 8.2\\
        $\text{granite-3.3-8b}$ & American Male & 12.7 & 98.9 & 40.5 & 97.0 & 84.1 & 68.5 & 22.1 & 10.3\\
        $\text{granite-3.3-8b}$ & British Female & 55.7 & 93.8 & 98.4 & 34.7 & 86.5 & 65.1 & 23.9 & 10.6\\
        $\text{granite-3.3-8b}$ & British Male & 74.3 & 84.5 & 98.2 & 34.7 & 86.1 & 66.7 & 26.3 & 10.5\\
        \bottomrule
    \end{tabular}
    \caption{Performance breakdown under \textbf{Linguistic Adversarial Attacks}. All values are reported in percentages \%.}
    \label{table:linguistic_attacks}
    \vspace{-10pt}
\end{table*}

Table \ref{table:linguistic_attacks} highlights the comparative stealth of linguistic attacks. A critical failure mode is observed in gemma-3n-E4B on the “American Female” profile; the attack achieves a 100\% Success Rate (ASR), yet the model maintains an exceptionally high coherence score ($\Phi_{\text{Coh}}^{\text{PER}} = 95.3\%$) and negligible dissonance ($\Psi_{\text{Diss}}^{\text{PER}} = 4.7\%$). This indicates that the linguistic perturbations decoupled the reasoning process from the acoustic evidence, resulting in high internal logical consistency despite an incorrect final verdict. In contrast, Qwen2-Audio demonstrates greater robustness to these textual perturbations, maintaining a lower ASR ($\approx 26\%$) and preserving the logical alignment between the generated transcript and the acoustic input.

\subsection{Reasoning-Verdict Coherence Under Attacks}

\renewcommand{\tabcolsep}{5pt}
\begin{table*}[tb!]
    \footnotesize
    \centering
    \begin{tabular}{lcccccccccccc}
        \toprule
        \textbf{ALMs} & \multicolumn{2}{c}{\textbf{Prosody}} & \multicolumn{2}{c}{\textbf{Disfluency}} & \multicolumn{2}{c}{\textbf{Speed}} & \multicolumn{2}{c}{\textbf{Speaking Style}} & \multicolumn{2}{c}{\textbf{Liveliness}} & \multicolumn{2}{c}{\textbf{Quality}}\\
& \textbf{ORG} & \textbf{PER} & \textbf{ORG} & \textbf{PER} & \textbf{ORG} & \textbf{PER} & \textbf{ORG} & \textbf{PER} & \textbf{ORG} & \textbf{PER} & \textbf{ORG} & \textbf{PER}\\
\cmidrule(lr){2-3} \cmidrule(lr){4-5} \cmidrule(lr){6-7} \cmidrule(lr){8-9} \cmidrule(lr){10-11} \cmidrule(lr){12-13}
Audio-Flamingo-3 & 89.80 & 65.25 & 70.06 & 69.12 & 75.31 & 84.61 & 71.19 & 78.45 & 89.63 & 85.07 & 83.25 & 88.35 \\
Qwen2-Audio-7B & 87.88 & 80.10 & 86.74 & 78.43 & 87.97 & 81.20 & 87.20 & 80.49 & 85.52 & 76.88 & 87.67 & 79.78 \\
Phi-4-multimodal & 90.14 & 72.72 & 88.80 & 71.95 & 89.94 & 72.29 & 90.00 & 73.25 & 88.05 & 69.66 & 87.65 & 70.69 \\
gemma-3n-E4B & 66.62 & 70.40 & 68.81 & 66.76 & 66.07 & 68.57 & 64.61 & 68.90 & 67.87 & 66.17 & 67.00 & 68.38 \\
granite-3.3-8b & 88.62 & 69.85 & 87.64 & 68.77 & 88.69 & 70.53 & 85.52 & 70.96 & 85.45 & 69.05 & 83.13 & 69.43 \\
\cmidrule(lr){1-13}
Average & 84.61 & 71.66 & 80.41 & 71.01 & 81.60 & 75.44 & 79.70 & 74.41 & 83.30 & 73.37 & 81.74 & 75.33 \\

        \bottomrule
    \end{tabular}
    \caption{$\Phi_{\text{Coh}}$ analysis, ORG is original reasoning, and PER is perturbed reasoning}
    \label{table:RQ2_coherence}
    \vspace{-10pt}
\end{table*}

Table \ref{table:RQ2_coherence} illustrates the dimension-wise erosion of logical consistency. We observe a systematic breakdown across all forensic axes, but the degradation is most severe in the foundational dimensions of \textit{Disfluency} and \textit{Prosody}, which suffer absolute drops of $11.52\%$ and $10.05\%$ respectively on average. This indicates that adversarial perturbations primarily disrupt the model's ability to articulate fine-grained acoustic properties; the model struggles to form a coherent logical chain regarding the speaker's rhythm and hesitation patterns when the signal is contaminated. Notably, the \textit{Quality} dimension also exhibits a sharp decline ($\sim9.3\%$), reflecting the model's confusion in distinguishing between environmental noise (a natural feature) and adversarial artifacts (an attack signature).

\subsection{Reasoning-Verdict Dissonance Under Attacks}

\renewcommand{\tabcolsep}{5pt}
\begin{table*}[tb!]
    \footnotesize
    \centering
    \begin{tabular}{lcccccccccccc}
        \toprule
        \textbf{ALMs} & \multicolumn{2}{c}{\textbf{Prosody}} & \multicolumn{2}{c}{\textbf{Disfluency}} & \multicolumn{2}{c}{\textbf{Speed}} & \multicolumn{2}{c}{\textbf{Speaking Style}} & \multicolumn{2}{c}{\textbf{Liveliness}} & \multicolumn{2}{c}{\textbf{Quality}}\\
& \textbf{ORG} & \textbf{PER} & \textbf{ORG} & \textbf{PER} & \textbf{ORG} & \textbf{PER} & \textbf{ORG} & \textbf{PER} & \textbf{ORG} & \textbf{PER} & \textbf{ORG} & \textbf{PER}\\
\cmidrule(lr){2-3} \cmidrule(lr){4-5} \cmidrule(lr){6-7} \cmidrule(lr){8-9} \cmidrule(lr){10-11} \cmidrule(lr){12-13}
Audio-Flamingo-3 & 35.36 & 39.91 & 21.15 & 34.49 & 12.97 & 15.84 & 18.87 & 23.37 & 19.27 & 17.13 & 16.22 & 12.39 \\
Qwen2-Audio-7B & 23.68 & 17.67 & 21.41 & 18.97 & 22.17 & 16.90 & 22.17 & 17.69 & 21.41 & 18.47 & 16.98 & 18.29 \\
Phi-4-multimodal & 18.93 & 29.37 & 28.58 & 34.83 & 26.73 & 34.31 & 33.36 & 34.60 & 23.16 & 35.61 & 30.18 & 38.56 \\
gemma-3n-E4B & 40.50 & 34.13 & 43.55 & 37.20 & 42.49 & 34.14 & 42.44 & 34.47 & 44.68 & 37.58 & 43.90 & 35.66 \\
granite-3.3-8b & 20.41 & 16.90 & 20.51 & 17.63 & 20.91 & 17.73 & 19.82 & 17.06 & 16.44 & 17.51 & 21.97 & 18.73 \\
\cmidrule(lr){1-13}
Average & 27.78 & 27.60 & 27.04 & 28.62 & 25.05 & 23.78 & 27.33 & 25.44 & 25.01 & 25.26 & 25.85 & 24.73 \\

        \bottomrule
    \end{tabular}
    \caption{$\Psi_{\text{Diss}}$ analysis, ORG is original reasoning, and PER is perturbed reasoning}
    \label{table:RQ3_dissonance}
    \vspace{-10pt}
\end{table*}

Table \ref{table:RQ3_dissonance} identifies the forensic dimensions most sensitive to adversarial perturbations. The \textit{Quality} dimension serves as a primary indicator of manipulation, particularly in the Phi-4-multimodal model, where dissonance increases from 38.56\% under adversarial conditions. This suggests that the reasoning layer generates evidence inconsistent with the final classification, even when the model fails to correctly categorize the audio. Conversely, dimensions such as \textit{Liveliness} exhibit heterogeneous behaviors: while Qwen2-Audio suppresses internal dissonance, smaller models such as Gemma demonstrate a reduction in dissonance under attack (e.g., Liveliness dissonance shifts from (44.68\% to 37.58\%). These results indicate that certain model architectures may lack the internal consistency required to flag adversarial inputs, potentially leading to logical alignment with incorrect classification outputs.

\subsection{Correlation Analysis of Perceptual Grounding and Reasoning Shifts}
\label{app:extended_corr}

\renewcommand{\tabcolsep}{5pt}
\begin{table*}[tb!]
\footnotesize
\centering
\begin{tabular}{lcccccccccccc}
\toprule
& \multicolumn{6}{c}{\textbf{Acoustic Attacks}} & \multicolumn{6}{c}{\textbf{Linguistic Attacks}} \\

\cmidrule(lr){2-7} \cmidrule(lr){8-13}
\textbf{Dimension} & $r_{\text{Coh}}$ & $p_{\text{Coh}}$ & $r_{\text{Diss}}$ & $p_{\text{Diss}}$ & $r_{\text{ASR}}$ & $p_{\text{ASR}}$ & $r_{\text{Coh}}$ & $p_{\text{Coh}}$ & $r_{\text{Diss}}$ & $p_{\text{Diss}}$ & $r_{\text{ASR}}$ & $p_{\text{ASR}}$ \\
\cmidrule(lr){2-7} \cmidrule(lr){8-13}

Prosody        & 0.274 & .323 & -0.566* & .028 & -0.199 & .478 & 0.156 & .511 & 0.602** & .005 & -0.142 & .551 \\
Disfluency     & -0.353 & .197 & -0.362 & .185 & -0.173 & .537 & -0.115 & .628 & -0.052 & .828 & -0.332 & .153 \\
Speed          & 0.049 & .863 & 0.643** & .010 & -0.407 & .133 & -0.727*** & .000 & -0.187 & .429 & 0.082 & .731 \\
Speaking Style & 0.558* & .031 & -0.127 & .652 & -0.261 & .347 & -0.121 & .611 & 0.541* & .014 & -0.043 & .857 \\
Liveliness     & 0.381 & .161 & -0.502 & .057 & 0.177 & .528 & 0.430 & .059 & 0.428 & .060 & -0.142 & .551 \\
Quality        & 0.475 & .073 & 0.220 & .430 & 0.316 & .251 & -0.097 & .683 & -0.262 & .264 & -0.395 & .085 \\
\bottomrule
\end{tabular}
\caption{Pearson correlation analysis of perceptual grounding ($\Phi_{\text{Perc}}$) against reasoning shifts (Coherence, Dissonance, and ASR). Values show $r$ (coefficient) and $p$ (p-value). * $p < 0.05$, ** $p < 0.01$, *** $p < 0.001$}
\label{tab:combined_corr}
\end{table*}

To characterize the factors driving reasoning instability, we computed the Pearson correlation ($r$) between foundational perceptual grounding ($\Phi_{\text{Perc}}$) and the resulting shifts in reasoning-verdict coherence ($\Delta\Phi_{\text{Coh}}$), dissonance ($\Delta\Psi_{\text{Diss}}$), and attack susceptibility ($\Delta\text{ASR}$) under adversarial conditions. The results, summarized in Table \ref{tab:combined_corr}, reveal that perceptual grounding is not uniformly predictive across threat models. Under acoustic perturbations, \textit{Speed} perception demonstrates a strong positive correlation with dissonance increase ($r=0.643, p=0.010$), indicating that grounding in temporal rhythm serves as an internal diagnostic signal for adversarial interference; when this feature is perceived, the model exhibits heightened internal conflict despite classification failures. Conversely, \textit{Speaking Style} serves as a stability anchor for coherence ($r=0.558, p=0.031$). Interestingly, \textit{Prosody} perception is negatively correlated with dissonance ($r=-0.566, p=0.027$), implying that models with high prosodic sensitivity may inadvertently suppress internal conflict detection, potentially leading to over-confident, albeit incorrect, classifications.

The linguistic regime reveals a distinct divergence in these dependencies. \textit{Speed} perception is strongly negatively correlated with coherence ($r=-0.727, p<0.001$), suggesting that rhythmic grounding is paradoxically detrimental when models face semantic-level manipulation, likely due to feature misalignment. Furthermore, significant positive correlations between \textit{Prosody} and dissonance ($r=0.602, p=0.005$), as well as \textit{Speaking Style} and dissonance ($r=0.541, p=0.014$), indicate that linguistic perturbations, which operate via semantic substitution, systematically induce logical friction in models sensitive to phonetic and stylistic nuances. These findings collectively demonstrate that reasoning-verdict dissonance serves as a valuable forensic artifact, surfacing adversarial presence through the model's inability to reconcile perturbed inputs with its learned internal logic.

\subsection{Linguistic Adversarial Attack Detection}

\renewcommand{\tabcolsep}{3pt}
\begin{table}[tb!]
    \footnotesize
    \centering
    
    \begin{tabular}{lcccc}
        \toprule
        & \textbf{Coef} & \textbf{dy/dx} & \textbf{std err} & \textbf{P>|z|} \\
        \cmidrule(lr){2-5}
        Prediction Score & 1.3703 & 0.2379 & 0.012 & 0.000 \\
        \cmidrule(lr){1-5}
        $\Phi_{\text{Coh}}$(Prosody) & -0.2306 & -0.0400 & 0.012 & 0.000 \\
        $\Phi_{\text{Coh}}$(Disfluency) & -0.0783 & -0.0136 & 0.013 & 0.000 \\
        $\Phi_{\text{Coh}}$(Speed) & -0.0032 & -0.0006 & 0.014 & 0.818 \\
        $\Phi_{\text{Coh}}$(Speaking Style) & 0.0185 & 0.0032 & 0.013 & 0.158 \\
        $\Phi_{\text{Coh}}$(Liveliness) & -0.1279 & -0.0222 & 0.013 & 0.000 \\
        $\Phi_{\text{Coh}}$(Quality) & -0.0157 & -0.0027 & 0.013 & 0.220 \\
        \cmidrule(lr){1-5}
        $\Phi_{\text{Perc}}$(Prosody) & 0.2452 & 0.0426 & 0.025 & 0.000 \\
        $\Phi_{\text{Perc}}$(Speaking Style) & -0.0308 & -0.0053 & 0.024 & 0.192 \\
        $\Phi_{\text{Perc}}$(Speed) & 0.3799 & 0.0660 & 0.013 & 0.000 \\
        \bottomrule
    \end{tabular}
    \caption{Logistic regression analysis for Linguistic attack detection.}
    \label{tab:linguistic_regression}
\end{table}

\begin{table}[tb!]
    \footnotesize
    \centering
    
    \begin{tabular}{lccc}
        \toprule
        \textbf{Model} & \textbf{Precision} & \textbf{Recall} & \textbf{F1 Score} \\ 
        \cmidrule(lr){2-4}
        LightGBM & 82.19\% & 92.64\% & 87.10\% \\
        XGBoost & 82.70\% & 91.13\% & 86.71\% \\
        Random Forest & 81.77\% & 84.34\% & 83.04\% \\
        AdaBoost & 81.60\% & 81.59\% & 81.60\% \\ 
        SVM (poly kernel) & 81.65\% & 72.86\% & 77.00\% \\
        Logistic Regression & 83.33\% & 66.96\% & 74.26\% \\
        \bottomrule
    \end{tabular}
    \caption{Detection performance metrics for Linguistic attacks.}
    \label{tab:linguistic_performance}
\end{table}

Table \ref{tab:linguistic_regression} presents the multivariate logistic regression analysis for the detection of linguistic adversarial attacks. The reasoning-based metrics, specifically $\Phi_{\text{Perc}}$(Speed) ($\beta = 0.3799, p < 0.001$) and $\Phi_{\text{Perc}}$(Prosody) ($\beta = 0.2452, p < 0.001$), demonstrate the highest predictive utility. These coefficients suggest that linguistic perturbations alter the model's perception of temporal and prosodic attributes, resulting in systematic variance in the generated reasoning traces. The performance metrics in Table \ref{tab:linguistic_performance} confirm that these reasoning-based features facilitate attack detection with an F1 score of 87.10\% using LightGBM, suggesting that reasoning traces contain sufficient information to identify linguistic manipulation even when standard classification headers do not.

\subsection{Acoustic Adversarial Attack Detection}

\renewcommand{\tabcolsep}{3pt}
\begin{table}[tb!]
    \footnotesize
    \centering
    \begin{tabular}{lcccc}
        \toprule
        & \textbf{Coef} & \textbf{dy/dx} & \textbf{std err} & \textbf{P>|z|} \\
        \cmidrule(lr){2-5}
        Prediction Score & -0.7799 & -0.1636 & 0.015 & 0.000 \\
        \cmidrule(lr){1-5}
        $\Phi_{\text{Coh}}$(Prosody) & -0.1203 & -0.0252 & 0.018 & 0.000 \\
        $\Phi_{\text{Coh}}$(Disfluency) & -0.1332 & -0.0279 & 0.016 & 0.000 \\
        $\Phi_{\text{Coh}}$(Speed) & -0.0719 & -0.0151 & 0.018 & 0.000 \\
        $\Phi_{\text{Coh}}$(Speaking Style) & -0.0345 & -0.0072 & 0.018 & 0.050 \\
        $\Phi_{\text{Coh}}$(Liveliness) & -0.1077 & -0.0226 & 0.018 & 0.000 \\
        $\Phi_{\text{Coh}}$(Quality) & -0.0714 & -0.0150 & 0.017 & 0.000 \\
        \cmidrule(lr){1-5}
        $\Phi_{\text{Perc}}$(Prosody) & -0.1729 & -0.0363 & 0.028 & 0.000 \\
        $\Phi_{\text{Perc}}$(Quality) & 0.1330 & 0.0279 & 0.015 & 0.000 \\
        $\Phi_{\text{Perc}}$(Speaking Style) & 0.3242 & 0.0680 & 0.027 & 0.000 \\
        \bottomrule
    \end{tabular}
    \caption{Logistic regression analysis for Acoustic attack detection.}
    \label{tab:acoustic_regression}
\end{table}

\begin{table}[tb!]
    \footnotesize
    \centering
    \begin{tabular}{lccc}
        \toprule
        \textbf{Model} & \textbf{Precision} & \textbf{Recall} & \textbf{F1 Score} \\ 
        \cmidrule(lr){2-4}
        LightGBM & 72.33\% & 82.26\% & 76.97\% \\
        XGBoost & 69.48\% & 81.62\% & 75.06\% \\
        Random Forest & 70.56\% & 78.41\% & 74.28\% \\
        SVM (poly kernel) & 62.19\% & 87.95\% & 72.86\% \\
        Logistic Regression & 62.06\% & 87.01\% & 72.45\% \\
        AdaBoost & 63.86\% & 79.61\% & 70.87\% \\
        \bottomrule
    \end{tabular}
    \caption{Detection performance metrics for Acoustic attacks.}
    \label{tab:acoustic_performance}
\end{table}

Table \ref{tab:acoustic_regression} summarizes the regression analysis for acoustic perturbations. Unlike linguistic attacks, acoustic perturbations are characterized by a negative correlation with reasoning consistency. Statistically significant coefficients are observed across multiple reasoning dimensions, most notably $\Phi_{\text{Coh}}$(Disfluency) ($\beta = -0.1332, p < 0.001$) and $\Phi_{\text{Coh}}$(Prosody) ($\beta = -0.1203, p < 0.001$). These negative associations indicate that acoustic noise reduces the model's capacity to maintain logical entailment between intermediate reasoning premises and the final classification. As demonstrated in Table \ref{tab:acoustic_performance}, these structural dependencies within the reasoning process provide the basis for detection, yielding an F1 score of 76.97\% for the LightGBM classifier.

\subsection{Generalization to External Benchmarks}

\begin{table}[tb!]
    \footnotesize
    \centering
    \renewcommand{\tabcolsep}{3pt}
    \begin{tabular}{lccc}
        \toprule
        & \textbf{ASV 2021} & \textbf{FOR} & \textbf{ITW}  \\
        \cmidrule(lr){2-4}

        $\text{Audio-Flamingo-3}^{NON}$ & 93.2\% & 73.5\% & 62.3\% \\
        $\text{Audio-Flamingo-3}^{RSN}$ & 93.8\% & 60.3\% & 40.0\% \\
        \cmidrule(lr){1-4}
        
        $\text{Qwen2-Audio-7B}^{NON}$ & 94.8\% & 52.7\% & 37.2\% \\
        $\text{Qwen2-Audio-7B}^{RSN}$ & 92.7\% & 54.3\% & 37.2\% \\
        
        \cmidrule(lr){1-4}
        $\text{granite-3.3-8b}^{NON}$ & 96.59\% & 53.81\% & 52.8\% \\
        $\text{granite-3.3-8b}^{RSN}$ & 94.57\% & 51.16\% & 43.6\% \\
        
        \cmidrule(lr){1-4}
        $\text{Phi-4-multimodal}^{NON}$ & 94.7\% & 51.2\% & 42.9\% \\
        $\text{Phi-4-multimodal}^{RSN}$ & 93.5\% & 60.2\% & 38.3\% \\
        
        \cmidrule(lr){1-4}
        $\text{gemma-3n-E4B}^{NON}$ & 98.0\% & 87.3\% & 84.5\% \\
        $\text{gemma-3n-E4B}^{RSN}$ & 64.2\% & 46.6\% & 55.4\% \\
    
        \bottomrule
    \end{tabular}
    \caption{Cross-validation accuracy (\%) of ALMs trained on ASVspoof 2019 and evaluated on ASVspoof 2021, Fake Or Real (FOR), and In The Wild (ITW) datasets.}
    \label{table:extend_dataset_result}
    \vspace{-10pt}
\end{table}

To evaluate the robustness of our ALM-based detection beyond the ASVspoof 2019 training distribution, we conducted cross-validation experiments on three external datasets: ASVspoof 2021, Fake Or Real (FOR), and In The Wild (ITW). The results, summarized in Table \ref{table:extend_dataset_result}, indicate varying degrees of transferability across architectures. 

Notably, gemma-3n-E4B$^{NON}$ demonstrates strong zero-shot performance across all benchmarks, yet experiences a significant performance collapse in $RSN$ mode when evaluated on ASVspoof 2021 (dropping from 98.0\% to 64.2\%). Conversely, models like Qwen2-Audio-7B maintain more consistent performance across $NON$ and $RSN$ configurations in external environments. These results highlight that while reasoning-capable ALMs offer distinct forensic advantages, their performance in real-world, out-of-domain scenarios remains sensitive to the underlying model architecture and the domain shift inherent in the evaluation data.

\section{Datasets and Prompts}
\label{app:dataset}

\begin{figure*}
    \centering
    \begin{tcolorbox}[
        enhanced,
        colback=gray!5,        
        colframe=gray!50,     
        arc=2pt,              
        boxrule=0.5pt,        
        width=\textwidth,     
        fontupper=\small\ttfamily, 
    ]
    You are an expert in speech analysis in audio deepfake detection. Please analyze the speech under these criteria, then provide the conclusion and clearly state the audio is \$fake\$ or \$real\$ (put your decision in \$\$). \\

    - Prosody: Analyze Tone, Inflections, Cadence, Pitch, Monotone, Raspy, Emotion,... \\
    - Disfluency: Analyze Pauses, Filler Words \\
    - Speed: Analyze Fast, Slow, Rushed \\
    - Speaking Style: Accent, List, Articulation, Specific Word Pronunciation \\
    - Liveliness: like do you hear the breathing, or robotic sound \\
    - Quality: analyze background noise, computer noise, microphone feedback noise,... \\
    - Conclusion: Based on above analysis, give your inductive reasoning and clearly state the audio is \$fake\$ or \$real\$ (put your decision in \$\$)
    \end{tcolorbox}
    \caption{Chain-of-Thoughts prompt with step-by-step instructions}
    \label{fig:cot_prompt}
\end{figure*}

\begin{figure*}
    \centering
    \begin{tcolorbox}[
        enhanced,
        colback=gray!5,        
        colframe=gray!50,     
        arc=2pt,              
        boxrule=0.5pt,        
        width=\textwidth,     
        fontupper=\small\ttfamily, 
    ]
    - Prosody: The prosody is quite unnatural. The pitch is consistently high, which is unusual for a human voice. There is a noticeable flatness in the delivery, which is often a sign of a voice synthesizer. The cadence is also quite unnatural, lacking the natural pauses and inflections that would be expected in a conversation.\\
- Disfluency: The speech is perfectly fluent, which is a significant red flag. There are no filler words or pauses, which are typical of human speech. This suggests that the speech was likely generated by a voice synthesizer.\\
- Speed: The speaking rate is quite slow, which is unusual for a human voice. This suggests that the speech was likely generated by a voice synthesizer.\\
- Speaking Style: The articulation is quite clear, which is unusual for a human voice. However, there are some noticeable errors in pronunciation, such as the incorrect use of "lending" (00:03) and the incorrect stress on "recently" (00:06). This suggests that the speech was likely generated by a voice synthesizer.\\
- Liveliness: There are no audible signs of life in the speech, such as breathing or other vocalizations. This suggests that the speech was likely generated by a voice synthesizer.\\
- Quality: The audio quality is quite low, with a noticeable amount of noise. This suggests that the speech was likely recorded with a low-quality microphone, which is common for voice synthesizers.\\
- Conclusion: Based on the analysis, the speech is likely \$fake\$. The audio quality is low, the speech is perfectly fluent and lacks any signs of life, and the prosody, disfluency, and speaking style are all indicative of a voice synthesizer.
    \end{tcolorbox}
    \caption{ASVspoof 2019, a reasoning training example for a \textbf{fake} audio}
    \label{fig:fake_example}
\end{figure*}

\begin{figure*}
    \centering
    \begin{tcolorbox}[
        enhanced,
        colback=gray!5,        
        colframe=gray!50,     
        arc=2pt,              
        boxrule=0.5pt,        
        width=\textwidth,     
        fontupper=\small\ttfamily, 
    ]
    - Prosody: The prosody of the speech is quite natural and consistent. There are no noticeable variations in pitch or intonation. The speech is delivered in a calm and steady manner, which is typical of a real person speaking in a neutral mood.\\
- Disfluency: The speech is quite fluent, with no noticeable hesitations or pauses. There are no fillers or pauses between words, which indicates that the speech was delivered smoothly without any interruptions. This is typical of a real person speaking in a calm and steady manner.\\
- Speed: The speech is delivered at a moderate pace, which is typical of a real person speaking in a neutral mood. There are no variations in speed, which indicates that the speech was delivered at a consistent pace.\\
- Speaking Style: The articulation is clear and precise, with no noticeable errors or mispronunciations. The accent is neutral and consistent, which indicates that the speaker is speaking in a neutral mood.\\
- Liveliness: There are no noticeable signs of liveliness or emotion in the speech. The voice is calm and steady, which is typical of a real person speaking in a neutral mood.\\
- Quality: The audio quality is quite good, with no noticeable background noise or distortion. The recording is clear and crisp, which indicates that the speech was recorded in a professional environment.\\
- Conclusion: Based on the analysis, the speech appears to be real. The prosody, disfluency, speed, speaking style, articulation, and quality are all consistent with a real person speaking in a neutral mood. There are no noticeable signs of artificiality or manipulation. \$real\$
    \end{tcolorbox}
    \caption{ASVspoof 2019, a reasoning training example for a \textbf{real} audio}
    \label{fig:real_example}
\end{figure*}

\subsection{Acoustic Perception: Dataset Construction}
\label{app:perception}

\begin{figure}[tb!]
    \centering
    \includegraphics[width=7.5cm]{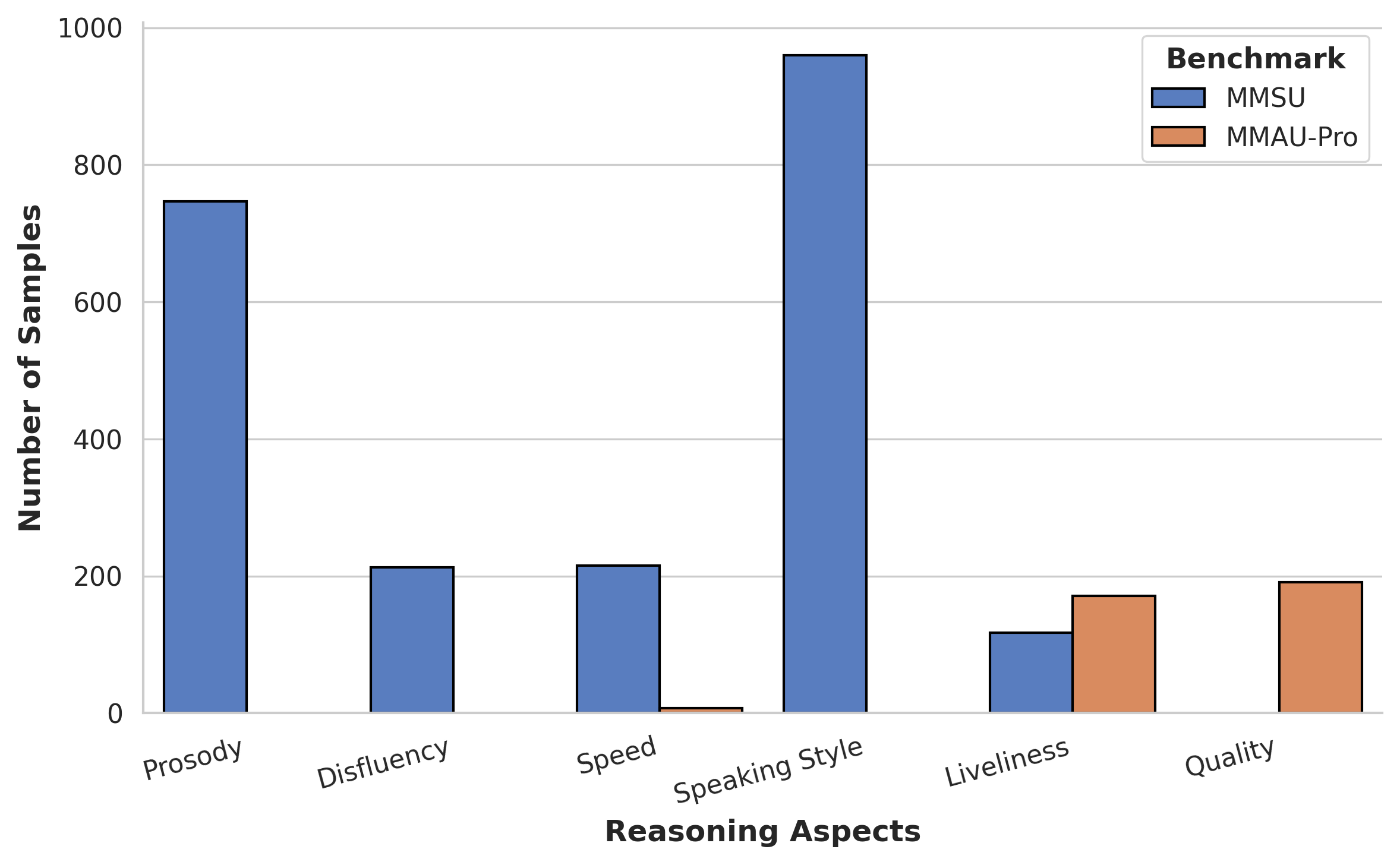}
    \caption{Distilled acoustic perception benchmark statistics from MMSU and MMAU-Pro.}
    \label{fig:distilled_rq1_benchmark_stats}
\end{figure}

To evaluate if the models accurately perceive basic audio details (RQ1), we curate a distilled benchmark, $\mathcal{D}_{Perc}$, by aggregating expert-annotated samples from two existing datasets: MMSU \cite{wang2026mmsu} and MMAU-Pro \cite{kumar2025mmauprochallengingcomprehensivebenchmark}. We organized these samples around six key reasoning aspects that are helpful for spotting deepfakes, such as prosody, disfluency, and audio quality.

The construction of $\mathcal{D}_{Perc}$ involves mapping specific sub-tasks from these source datasets to our six defined forensic dimensions. The final composition of the benchmark is categorized as follows:

\begin{itemize}[leftmargin=*,itemsep=2pt,parsep=0pt]
    \item \textbf{Prosody}: Distilled from MMSU, encompassing emotion recognition, intonation and pitch comparison, vocal range, speech stress, prolonged sounds, and volume comparison.
    \item \textbf{Disfluency}: Sourced from MMSU, focusing on disfluency detection and pause perception.
    \item \textbf{Speed}: Aggregated from MMSU (speed comparison, speech duration estimation) and MMAU-Pro (rhythmic pattern analysis).
    \item \textbf{Speaking Style}: Sourced from MMSU, including accent identification, consonant/vowel and plosive sound perception, syllable/near-homophone perception, speech act classification, and speaker demographic/identity prediction.
    \item \textbf{Liveliness}: Combined from MMSU (non-verbal sound detection) and MMAU-Pro (paralinguistic and emotion recognition).
    \item \textbf{Quality}: Derived from MMAU-Pro, targeting audio quality, artifacts and channel characteristics, spatial speech perception, and auditory source separation.
\end{itemize}

\paragraph{Dataset Statistics.}

In Figure \ref{fig:distilled_rq1_benchmark_stats}, $\mathcal{D}_{Perc}$ comprises $N=2,621$ samples with an overall average audio length of $7.89$ seconds. This includes $2,252$ samples from MMSU (average duration: $6.40$s) and $369$ samples from the MMAU-Pro subset (average duration: $17.02$s).

\subsection{Author-Based Human Annotations}
\label{sec:author_annotations}

To ensure the quality and consistency of our evaluations, all human-labeled data referenced in Section \ref{subsec:data_synthesis} (Reasoning Data Synthesis) and Appendix~\ref{app:imp_entailment} (Implementation Details of The Entailment Function) were annotated directly by the authors. This annotation effort comprises two main components:
\begin{itemize}
    \item \textbf{Human-in-the-Loop Seeds (Section~6.2):} The 50 real and 50 fake reasoning samples utilized as part of the seed data during the iterative refinement process for our dataset generation.
    \item \textbf{Evaluator Validation Set (Appendix~\ref{app:imp_entailment}:} The held-out set of 100 human-annotated reasoning traces used to evaluate the semantic stance classification of our fine-tuned Qwen3-8B entailment evaluator.
\end{itemize}

All annotations were conducted manually by the authors in strict accordance with the six forensic dimensions defined in our taxonomy (Prosody, Disfluency, Speed, Speaking Style, Liveliness, and Quality). No external crowdsourcing platforms or third-party annotators were employed, ensuring ethical data management and a consistent, rigorous labeling standard throughout the study.

\section{AI Assistants Usages}
\label{sec:ai_usage}
 During the preparation of this work, we utilized several generative AI assistants to support the research pipeline and manuscript development. Specifically, we employed $Gemini\text{-}3.1\text{-}pro$ to execute the $\textit{Cold Start}$ data synthesis process, generating the initial reasoning traces and the question bank $\mathcal{Q}_k$. For the manuscript preparation, $Gemini\text{-}3.1$ was used to assist with grammar checking, stylistic refinement, and the selection of precise word choices. 

We have rigorously reviewed and edited all AI-assisted outputs to ensure scientific accuracy. We maintain full responsibility for the final content of this publication.

\end{document}